\documentclass[conference]{IEEEtran}
\IEEEoverridecommandlockouts
\usepackage[backend=biber,style=ieee,natbib=true,maxbibnames=99]{biblatex} 
\renewcommand{\bibfont}{\footnotesize} % for IEEE bibfont size
\usepackage{amsmath,amssymb,amsfonts}
\usepackage{algorithmic}
\usepackage{textcomp}
\usepackage{xcolor}
\usepackage{booktabs}
\usepackage{url}
\usepackage{hyperref}
\usepackage{xspace}
\usepackage{microtype}
\usepackage{hyperref}
\usepackage{url}
\usepackage{booktabs}
\usepackage{rotating}
\usepackage{float}
\usepackage{url}
\usepackage{pdfpages}

\usepackage{lineno}
\usepackage{url}
\usepackage{bbm}
\usepackage{graphicx}
\usepackage{multirow}
\usepackage[noend]{algorithm2e}
\usepackage{enumitem}
\usepackage{bbm}
\usepackage{listings}
\definecolor{darkblue}{rgb}{0, 0, 0.5}
\hypersetup{colorlinks=true, citecolor=darkblue, linkcolor=darkblue, urlcolor=darkblue}
\usepackage[font=small,labelfont=bf]{caption}

\usepackage{tcolorbox}
\tcbuselibrary{many}
\tcbset{
    enhanced,
    breakable,
    attach boxed title to top left={
        xshift=0.5cm,
        yshift=-\tcboxedtitleheight/2
    },
    top=4mm,
    coltitle=black,
    colback=violet!20,
    colbacktitle=violet!60,
    beforeafter skip=\baselineskip,
}
\newenvironment{mybox}[1]{%
    \begin{tcolorbox}[title={#1}]%
    }{
    \end{tcolorbox}
}

\newcommand{\ft}{\texttt{FT\;+\;Merge}\xspace}
\newcommand{\ct}{\texttt{Central\;FT}\xspace}
\newcommand{\tall}{\texttt{TALL-masks}\xspace}
\newcommand{\emr}{\texttt{EMR-merging}\xspace}
\newcommand{\ties}{\texttt{TIES-merging}\xspace}
\newcommand{\ours}{\texttt{SIFT-Masks}\xspace}
\newcommand{\stitch}{\texttt{Loc.-and-Stitch}\xspace}

\def\BibTeX{{\rm B\kern-.05em{\sc i\kern-.025em b}\kern-.08em
    T\kern-.1667em\lower.7ex\hbox{E}\kern-.125emX}}
{\endlist}
\begin{document}

\title{Exact Unlearning of Finetuning Data \\ via Model Merging at Scale}

% \title{Title \\
% % {\footnotesize \textsuperscript{*}Note: Sub-titles are not captured for https://ieeexplore.ieee.org  and
% % should not be used}
% \thanks{Identify applicable funding agency here. If none, delete this.}
% }

% \author{Authors omitted for submission}
\author{
\IEEEauthorblockN{
Kevin Kuo\thanks{Direct correspondence to: kkuo2@andrew.cmu.edu},
Amrith Setlur,
Kartik Srinivas,
Aditi Raghunathan,
and Virginia Smith
}
% \IEEEauthorblockA{\textit{dept. name of organization (of Aff.)} \\
\textit{Carnegie Mellon University}\\
Pittsburgh, PA \\
}

\maketitle

\begin{abstract}
% Approximate unlearning has gained popularity as an approach to efficiently update a model so it (roughly) behaves as if it was not trained on a subset of data. However, in practice, approximate unlearning methods have been shown to be quite brittle and can easily be attacked to reveal supposedly unlearned information. 
Approximate unlearning has gained popularity as an approach to efficiently update an LLM so that it  behaves (roughly) as if it was not trained on a subset of data to begin with. However, recent works have shown that approximate unlearning is brittle in practice and can easily be attacked to reveal supposedly unlearned information. To alleviate issues with approximate unlearning, we instead propose \ours~(\textbf{SI}gn-\textbf{F}ixed \textbf{T}uning-Masks), a method for one-shot finetuning and model merging that enables \emph{exact unlearning} at scale. \ours~addresses two key limitations of standard model merging: (1) merging a large number of tasks can severely harm utility; and (2) methods that boost utility by sharing extra information across tasks make exact unlearning prohibitively expensive. \ours~solves these issues by (1) applying local masks to recover task-specific performance; and (2) constraining finetuning to align with a global sign vector as a lightweight approach to  determine masks independently before merging. Across four settings where we merge up to 2,500 models, \ours~improves accuracy by 5-80\% over na\"ive merging and uses up to 250$\times$ less compute for exact unlearning compared to other merging baselines.
\end{abstract}


\section{Introduction}
\label{sec:intro}

Modern machine learning applications often require finetuning a pretrained model on a collection of data. However, once a model has been finetuned, it may be desirable to \emph{unlearn}  a subset of data and produce a model identical to one trained as if the data were never present. This is because finetuning data can introduce risks such as harmful knowledge or private information~\citep{carliniquantifying,more2024towards,su2024extracting,ahmadian2024mix}. Moreover, data privacy regulations such as GDPR and CCPA state that consumers have a ``right to be forgotten'' (RTBF)~\citep{protection2018general}, which requires data controllers to remove personal data in accordance with deletion requests. However, the exact requirements of RTBF are unclear for ML models, which are derivatives of personal data~\cite{juliussen2023algorithms}. To address these concerns, there has been significant interest in methods for \emph{machine unlearning} that can efficiently remove the influence of data from a model~\citep{cao2015towards,ginart2019making,tarun2023fast}. However, existing methods face key limitations: heuristic approximate unlearning methods lack guarantees---leading to exposure of supposedly unlearned information~\citep{hu2024jogging,lucki2024adversarial,deeb2024unlearning}, while exact unlearning methods have prohibitively expensive relearning costs~\citep{bourtoule2021machine,yan2022arcane,chen2022recommendation,li2024making,chowdhury2024towards}.

These two limitations raise a key question: can we change the way models are designed or finetuned so that an \emph{efficient exact unlearning method exists}? In this work, we propose using \emph{model merging} \color{red}\cite{ilharcoediting} \color{black} to enable both exact and efficient unlearning. First, given a dataset split over several tasks (\textit{e.g.}, a set of clients whose data we may wish to unlearn), we first finetune (\texttt{FT}) a pretrained model separately on each task to obtain a set of \emph{local} models. 
% While it is easy to unlearn a task by discarding its local model, this framework is limited by both high storage costs and lack of collaboration across tasks. Therefore, 
Next,
we average (\texttt{Merge}) the local models' weights to produce a single \emph{merged} model and then discard the local models. To unlearn a particular task, we can simply retrain
% (with deterministic data sampling)
the local model for that task and \emph{unmerge} (subtract) it from the merged model. This results in a model where the task has been unlearned exactly, \textit{i.e.}, it matches the merged model from the remaining tasks.

% \begin{wrapfigure}{r}{0.4\textwidth}
\begin{figure}
    \vspace{-0.65cm}
    \centering
    \includegraphics[trim={0 0.3cm 0 0},clip,width=1.0\linewidth]{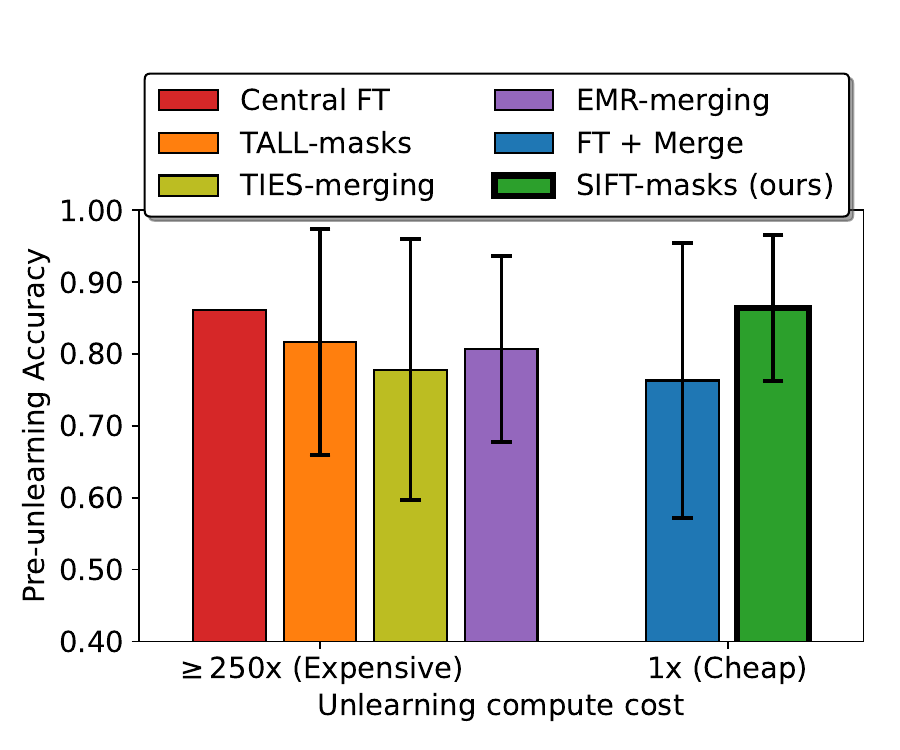}
     \vspace{-0.4cm}
    \caption{\textbf{We finetune 500 tasks (corresponding to  user tweets) on Sent140~\cite{go2009twitter}, and report the accuracy after finetuning (y-axis) as well as the cost to exactly unlearn all tasks (x-axis).} To enable efficient exact unlearning without sacrificing accuracy, we propose \ours, a method for one-shot finetuning and model merging that can match central training accuracy (\ct) while being as efficient for exact unlearning as na\"ive merging (\ft).}
    \label{fig:sift-intro}
    \vspace{-0.45cm}
\end{figure}
% \end{wrapfigure}

Finetuning and merging of this form (\ft) offers two key advantages over standard baselines for exact unlearning, such as model ensembling (e.g., output averaging)~\cite{bourtoule2021machine} or finetuning over a combined central dataset (\ct). First, it is cheaper to store a single merged model compared to a ensemble of local models, as the storage cost of these local models scales with the number of tasks (see Table~\ref{tab:compare_methods}). Second, unlearning requires only a small amount of compute to finetune over a single forget task, while \ct~requires re-finetuning over all retained tasks. Finally, \ft~allows for updating existing tasks or including entirely new tasks in a modular way while preserving exactness.

\begin{table*}
\normalsize
\centering
\begin{tabular}{lccc}
\toprule
\textbf{Method} & \textbf{Description} & \textbf{Unlearning Cost} & \textbf{Storage}\\
\midrule
\ct
& Train a single model on all pooled data & $T$ & $M$\\
\midrule
\multicolumn{2}{l}{\textit{\textbf{(1) Exact unlearning methods}}} \\
\midrule
\texttt{SISA} \citep{bourtoule2021machine} & Ensemble models trained on random data shards & $T/S$ & $MS$\\
\midrule
\texttt{APA} \citep{hu2024exact} & Merge models trained on data-driven shards & $T$ & $MS$\\
\midrule
\multicolumn{2}{l}{\textit{\textbf{(2) Model merging methods}}} \\
\midrule
\ft~\citep{ilharcoediting} & Merge models trained on task-level shards & $1$ & $M$\\
\midrule
\ties~\citep{yadav2024ties} & Elects a global sign vector before merging & $T$ & $M$\\
\midrule
 \stitch~\citep{he2024localize} & Sparsifies models before merging & $1$ & $M$ \\
\midrule
\multicolumn{2}{l}{\textit{\textbf{(3) Merging-and-localization methods}}} \\
\midrule
\tall~\citep{wang2024localizing} & \multirow{2}{*}{Construct masks using merged and local models} & \multirow{2}{*}{$T$} & \multirow{2}{*}{$M(1 + \frac{T}{32})$} \\
\emr~\citep{huang2024emr} & & & \\

\midrule
\ours~(this work) & Construct masks using only local models & $1$ & $M(1 + \frac{T}{32})$ \\
\bottomrule
\end{tabular}
\caption{\textbf{Comparison of various model merging and ensembling methods which can be applied to exact unlearning.} First, \ct~na\"ively trains a new model on all retain tasks, which is prohibitive in terms of compute. Prior \textbf{exact unlearning methods (1)} based on merging and ensembling reduce unlearning costs by modifying how data is partitioned into tasks. However, these methods store all local models and only ensemble them at inference, which has high storage costs. \textbf{Model merging methods (2)} usually do not consider exact unlearning and degrade as the number of tasks grows. Finally, our work focuses on \textbf{merging-and-localization methods (3)}. While \ft~is a cheap baseline, improvements to this method do not consider exact unlearning and thus increase unlearning costs by a factor of $T$. $T$ is the number of retained tasks, $S$ is the number of shards (disjoint partition of tasks), and $M$ is the number of model parameters. ``Unlearning Cost'' is the worst-case compute cost of unlearning a single task in terms of the number of tasks we must retrain the model on. 
}
\label{tab:compare_methods}
\end{table*}
As shown in Figure~\ref{fig:sift-intro}, finetuning using the one-shot finetuning and merging strategy indeed leads to  efficient exact unlearning: unlearning 500 tasks using \ft is $250 \times$ cheaper than performing exact unlearning by retraining a model on the revised data (\texttt{Central FT}). However, performing \ft with simple averaging can significantly degrade accuracy, particularly when merging a large number of models.  It is thus natural to consider whether more sophisticated merging approaches could boost accuracy while keeping unlearning costs low.

% {Unfortunately}, as we show in Figure~\ref{fig:sift-intro}, performing \ft with simple averaging can significantly degrade accuracy, particularly when merging a large number of models. Still, merging is indeed a cheap way to enable exact unlearning; unlearning all 500 tasks in this experiment using \ft is $250 \times$ cheaper than performing exact unlearning by retraining a model on all the revised data (\texttt{Central FT}).  It is thus natural to consider whether more sophisticated merging approaches could boost accuracy while keeping unlearning costs low.

{Unfortunately}, state-of-the-art merging methods which improve accuracy at scale like \texttt{EMR-merging}~\citep{huang2024emr} or \texttt{TALL-masks}~\citep{wang2024localizing} rely on \textit{localization}---a method to recover performance by applying task-specific masks to the merged model. As shown in Figure~\ref{fig:sift-intro}, while these methods improve accuracy, they come at a unlearning high cost: current localization methods must relearn {all} local models to reconstruct local masks after a task is removed, rendering them as expensive (if not more so) than the na\"ive \texttt{Central\;FT} baseline.

To address this challenge, we propose \ours~(\textbf{SI}gn-\textbf{F}ixed \textbf{T}uning Masks), a lightweight model merging method that is uniquely suited for large-scale unlearning of finetuning data. \ours~performs merging by initializing a global random sign vector and constraining the entries within a task vector to agree with this vector while setting the others to zero.
% After applying the local masks, the sign of each dimension will match across all tasks model parameters, which improves accuracy by reducing sign conflicts when the local models are merged.
Notably, the global sign vector is chosen independently of the tasks and their data (\textit{i.e.}, before any training begins).
% Thus the mask used to localize the merged model depends only on the local model,  (and not the merged model). This naturally produces a set of local masks without sharing any information across tasks,
This design choice substantially reduces the cost of unlearning,  matching that of the efficient na\"ive averaging baseline, while  also improving the accuracy of the merged model relative to simple averaging. 
In summary, our contributions are as follows:

\begin{enumerate}[leftmargin=*]
\item We consider techniques for large-scale finetuning that can enable \textit{exact unlearning} by finetuning over tasks in a one-shot manner and then merging the resulting models.
\item We merge up to 2,500 models---greater than two orders of magnitude more tasks than prior work on model merging~\citep{ilharcoediting,wang2024localizing}. This mirrors more realistic settings of machine unlearning where models are finetuned over a large set of tasks (e.g., a set of clients or collection of documents), and tasks contribute subtly differing data for a common learning objective rather than the tasks being largely disjoint.
\item We identify a key deficiency in simply applying recent, localization-based merging methods to our setting: current localization methods boost accuracy when merging by sharing extra information across tasks, which can in turn make exact unlearning computationally infeasible. 
\item We instead propose \ours, a finetuning method that makes model merging and localization computationally feasible for exact unlearning.
Unlike existing methods where masks require global information, \ours~obtains masks using only a random sign vector and local data. This allows us to maintain accuracy while enabling efficient exact unlearning at scale: As we show through extensive experiments on unlearning tasks, \ours~improves accuracy by 5-80\% over na\"ive merging while using up to 250$\times$ less compute for unlearning compared to other merging baselines.
\end{enumerate}
\section{Related Work}
\label{sec:relwork}

\textbf{Machine Unlearning.} Machine unlearning considers the problem of removing the influence of specific subsets of training data from a model after it has been trained~\cite{ginart2019making,bourtoule2021machine}. Unlearning applications often aim to perform unlearning over a large number of underlying tasks, such as users who wish to opt out of data sharing or data sources which are found to contain harmful information~\citep{li2024wmdp,jin2024rwku,maini2024tofu}. Therefore, it is important to design methods that can efficiently update a model in response to multiple deletion requests. 
Standard approaches for \emph{exact} unlearning---which provably remove the unlearning data and exactly match the model retrained from scratch (e.g. in terms of model parameters or training data sequence)---tend to have high computational costs from retraining over a large retain set or high storage costs from maintaining and ensembling models trained on disjoint shards of data~\citep{bourtoule2021machine,yan2022arcane,chen2022recommendation,li2024making,chowdhury2024towards}. On the other hand, \emph{approximate} unlearning methods do not guarantee an exact match with the retrained model and can be categorized as either \textit{heuristic} or \textit{provable}. \textit{Heuristic approximate} unlearning methods empirically approximate the outputs of an unlearned model~\citep{eldan2024whos,liu2024model}. Consequently, many prior works show that these approaches are brittle and can be easily attacked or probed to reveal supposedly unlearned information~\citep{marchant2022hard,bertran2024reconstruction,hu2024learn,hu2024jogging,bourtoule2021machine,tarun2023fast}. Meanwhile, \textit{provable approximate} unlearning methods use a probabilistic relaxation of the definition of exact unlearning~\cite{neel2021descent,wei2025provable} (see A.2 of \cite{ginart2019making}). However, these guarantees only hold in specific settings for finetuning (e.g. convex models), and it is not immediately clear how to extend them to full-parameter LLM finetuning. 
% Therefore, due to the challenges of evaluating approximate unlearning methods and/or their restrictive assumptions on the model, our work focuses on exact unlearning methods. 
 
% Unlearning is also a natural problem in distributed or federated settings where users share their data or model parameters; methods tailored to these settings can similarly be categorized as exact~\citep{qiu2023fedcio,xiong2023exact,xia2024edge} or approximate~\citep{wu2022federated,halimi2022federated}. 

\textbf{Model Ensembling.} Given the pitfalls of heuristic approximate unlearning, several approaches have emerged that modify the training or finetuning procedure to enable exact unlearning. Ensembles and mixture-of-experts are two common approaches that can make unlearning easier by training multiple base models over subsets of the training data and then combining the outputs. Several works in exact unlearning leverage ensembles to trade off storage and relearning costs~\cite{bourtoule2021machine,yan2022arcane,chen2022recommendation,li2024making,chowdhury2024towards}. Prior work has also studied unlearning with mixture-of-experts in both exact \cite{shi2025flexolmo} and heuristic approximate \cite{zhuang2025seuf} settings. However, for both approaches, the cost of maintaining models can become prohibitively expensive with a large number of tasks (e.g., hundreds to thousands of clients or documents), as considered in this work. We thus instead focus on model merging, which produces a single model of the same dimension as the base model weights to reduce storage and inference costs. 

%\textbf{Model Ensembling.} Ensembles and mixture-of-experts are the two common approaches to combine multiple base models into a single architecture by averaging their outputs. This modular structure poses unique questions for unlearning.  As mentioned above, several works in exact unlearning leverage ensembles to trade off storage and relearning costs~\cite{bourtoule2021machine,yan2022arcane,chen2022recommendation,li2024making,chowdhury2024towards}. Prior work has also studied unlearning with mixture-of-experts in both exact \cite{shi2025flexolmo} and \color{red}heuristic \color{black} approximate \cite{zhuang2025seuf} settings. In contrast to these works, we focus on model merging, which only runs one set of base model weights at inference time. \gs{I think we should more clearly explain why we didn't compare to ensembling approaches / what the downsides are}

\textbf{Model Merging.} Model (weight) merging is a natural alternative to ensembling that could be used to enable exact unlearning. However, it is unclear how to retain accuracy when merging a large number of models and whether merging approaches can be made efficient for exact unlearning. Current works from the model merging literature merge relatively few tasks and focus more on the benefits of scaling the model size~\citep{yadav2024matters}. Further, while some model merging works have considered unlearning, they only explore an heuristic approximate unlearning setting where the goal is to remove pretrained knowledge from the model. Specifically, \cite{ilharcoediting,kim2024negmerge} apply a negated task vector that roughly approximates gradient ascent on the pretrained model, while \cite{kadhe2024split} considers merging multiple heuristic approximate unlearned models. In contrast, our work focuses on exact unlearning of knowledge acquired during finetuning.

To explore the effectiveness of approaches for model merging in the context of exact unlearning, we consider both standard baselines and state-of-the-art approaches from the merging literature. Early work in model merging averages the parameters of multiple models trained with different hyperparameters on the same data to improve generalization~\citep{wortsman2022model}. Concurrent works extend this method to multi-task learning by training vision-language models on diverse vision tasks and then merging their parameters; these works study how to weight the average to improve model performance~\citep{matena2022merging,dimitriadis2023pareto,ilharcoediting}. Since then, many methods have been proposed to improve the quality of naive merging (\ft), such as modifying the finetuning method~\citep{ortiz2024task,yoshida2025mastering}, sparsifying task vectors~\citep{marczak2024magmax,yu2024language,he2024localize,davari2025model,zhang2025lori}, or selectively merging subsets of weights~\citep{ainsworth2022git,stoica2023zipit,ye2023merging,xu2024training}. However, adding complexity to \ft~can make it unsuitable for exact unlearning. For example, \ties~is similar in spirit to our work, as it seeks to minimize sign conflicts in \ft~\citep{yadav2024ties}. A crucial step of \ties~is \emph{electing} a global sign based on all models before merging occurs, but this dependence between models prior to merging makes exact unlearning non-trivial.

\textbf{Model Localization.} To improve the performance of model merging,  serveral approaches consider the use of \textit{localization}, which applies task-specific masks to recover performance lost during  merging~\citep{wang2024localizing,huang2024emr}. The storage benefit of localization methods over ensembling (storing all local models) is that each mask costs 1/32 of the storage of a model with the same number of 32-bit floating-point weights, and only one copy of 32-bit weights (the merged weights) need to be stored. In general, there are many techniques that use extra storage to preserve task-specific models. Beyond localization, other similar works try to store multiple but fewer models than the original number of local models~\citep{zhang2024channel,hu2024exact,lu2024twin}. However, all of these works are limited in scale: \cite{wang2024localizing} and \cite{huang2024emr} merge up to 30 models and all the other works we are aware of only merge up to 8 models.

\begin{figure*}[t!]
    % \vspace{-0.4cm}
    \centering
    \includegraphics[width=1.0\linewidth]{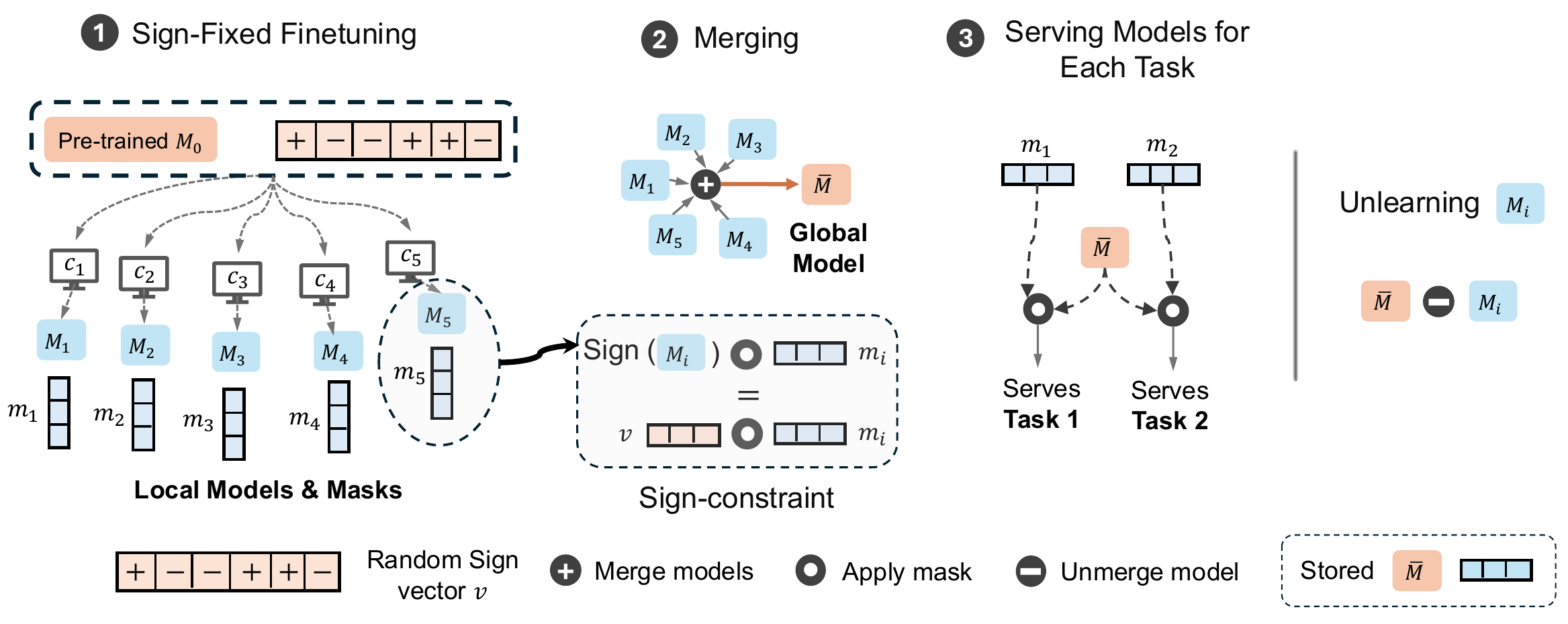}
    % \vspace{-0.2cm}
    \caption{\textbf{Overview of our method.} \ours consists of three steps: finetuning, merging, and localization. We start by running (1) \textbf{SI}gn-\textbf{F}ixed (Fine)-\textbf{T}uning, which initializes a random sign vector $v$ and constrains finetuning to match this sign vector or otherwise be sparse, producing a sparse local model $M_i$ with mask $m_i$. We then (2) merge these local models into a global model and only keep the masks $m_i$. When (3) serving task $c_i$, we apply $m_i$ to the merged model. Finally, to unlearn task $c_i$, we simply unmerge $M_i$ from $\overline{M}$ and discard $m_i$.}
    \label{fig:sift-figure}
    % \vspace{-0.2cm}
\end{figure*}

We critically note that subtle design choices made by a few methods in Table~\ref{tab:compare_methods} can result in unlearning costs similar to na\"ive central training. APA~\citep{hu2024exact} proposes \emph{data-driven clustering}, while \tall~\citep{wang2024localizing} and \emr~\citep{huang2024emr} propose \emph{masks which depend on the merged model}. Due to the extra information these methods share across tasks, subtracting model weights is insufficient to remove the influence of a task. For example, removing a client from a cluster may change the clusters output by the clustering algorithm, making it necessary to re-run the entire learning algorithm. In our proposed method (described below), we instead develop a localization procedure that can be used to improve utility but avoids the need to share additional information between tasks.

\section{\ours: Sign-Fixed Tuning and Masking}
\label{sec:method}
% \vspace{-0.2cm}

In this work we focus on developing a finetuning procedure that can enable \emph{exact} unlearning, having the benefit of provable unlearning by design. Given a dataset composed of a collection of tasks $c_1, ..., c_T$, a learning algorithm $\mathcal{A}$, and model $M = \mathcal{A}(\bigcup_{i\in [T]} c_i)$, \textit{exact unlearning} of task $c_u$ requires constructing a model $M_u = \mathcal{A}(\bigcup_{i\in [T]\setminus \{u\}} c_i)$ that matches the output from running $\mathcal{A}$ as if $c_u$ never existed. To avoid directly running $\mathcal{A}$ from scratch, efficient exact unlearning requires two key components: modifying $\mathcal{A}$ and designing an unlearning algorithm $\mathcal{U}$ such that $\mathcal{U}(M, c_u) = M_u$.

An important assumption in this definition is that the finetuning data $c_1, ..., c_T$ must be available to both 1) train $M$ by applying the modified $\mathcal{A}$ and 2) unlearn $c_u$ by applying $\mathcal{U}$. Therefore, exact unlearning cannot be applied when 1) the starting model is trained by a method other than $\mathcal{A}$ e.g. a pretrained model or 2) the data $c_u$ we want to unlearn is no longer available.

\textbf{Model merging provides both efficient and exact unlearning by design.} A key insight we have in this work is that model merging is an approach that, when used with one-shot finetuning, can produce a model that is amenable to efficient exact unlearning. We specifically consider a scenario where we finetune a pretrained model $M_0$ separately on several tasks, construct residual \emph{task vectors} $\tau_c = M_0 - M_c$, and then combine these to produce a multi-task vector $\overline{\tau} = \sum_{c\in[T]} \tau_c$ and a merged model $\overline{M} = M_0 + \overline{\tau}$~\citep{ilharcoediting}. Using these steps as the learning algorithm $\mathcal{A}$, we can unlearn task $c_u$ by simply subtracting its task vector, which exactly yields $\overline{\tau} - \tau_u = \sum_{c\in [T] \setminus \{u\}} \tau_c$, the merged model as if $c_u$ were never present. To avoid storing all the task vectors, we only store $\overline{\tau}$ and retrain $\tau_u$ when a deletion request is made. This retraining must be deterministic in order to reproduce the same weights of $\tau_u$. Below, we discuss two categories of randomness when training models with gradient descent:
\begin{enumerate}
    \item \textbf{Algorithmic randomness.} Higher-level functions like model initialization and data sampling can be controlled by fixing an RNG seed. Lower-level operations can also be made deterministic (e.g. by using deterministic CUDA kernels).
    \item \textbf{Software and hardware environment.} In order to ensure reproducibility, model retraining must be done in an identical environment. This would be a reasonable requirement for organizations that are maintaining the private data, which are likely to also have control of their computing infrastructure. 
\end{enumerate}

\begin{figure}
    % \vspace{-0.2cm}
    \centering
        \includegraphics[width=1\linewidth]{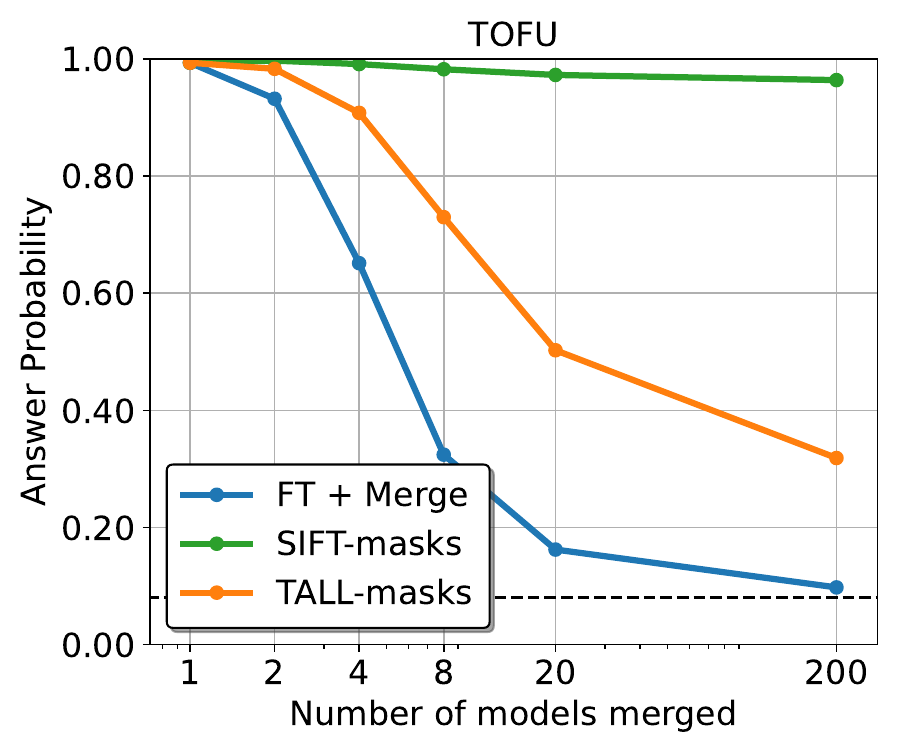}
    % \vspace{-0.2cm}
    \caption{\textbf{We finetune and merge models on up to 200 tasks (corresponding to TOFU authors)}. Na\"ive merging (\ft, $x>1$) degrades performance compared to applying local models ($x=1$); this issue becomes more severe as the number of models increases, potentially reducing performance to zeroshot quality. Our method \ours modifies finetuning to produce a merged model and a local mask for each task which can be applied to the merged model to recover task-specific performance.}
    \label{fig:merging_intro}
    % \vspace{-0.2cm}
\end{figure}

\textbf{Localization recovers utility lost from merging.} Prior work has shown that a merged model $\overline{M}$ tends to perform worse than the task-level models. In Figure \ref{fig:merging_intro}, we show that as the number of models increases, performance can degrade even further. \textbf{Localization (short for merging-and-localization)} is a promising approach to recover this lost performance; localization-based methods learn an additional mask $m_t$ for each task which approximates the local model weights once applied to the multi-task vector $M_t\approx M_0+m_t\odot\overline{\tau}$. We compare to two localization baselines: \tall, which merges by averaging and localizes by using a similarity threshold hyperparameter $\lambda_t$: $m_t = \mathbbm{1}\{|\tau_t| \geq |\overline{\tau} - \tau_t|\cdot\lambda_t\}$~\citep{wang2024localizing}; and \emr, which merges by taking a maximum over all weights which align with the average sign, localizes with sign agreement $m_t = \mathbbm{1}\{\tau_t \odot \overline{\tau} > 0\}$, and then rescales $m \odot \overline \tau$ to match the $\ell_1$ norm of $\tau_t$~\citep{huang2024emr}.

\textbf{Existing localization methods are unsuitable for exact unlearning.} Despite the effectiveness of localization in terms of improving accuracy, a critical limitation of existing methods is that masks depend on both the local and merged models. To satisfy exact unlearning, the local masks cannot directly be reused after an unlearning request is made and instead must be reconstructed using the new merged model. Furthermore, in order to reconstruct the masks, each local model has to be retrained, since we do not store the local models. Therefore, naively applying these methods for exact unlearning requires finetuning over the entire dataset after each deletion request, which is computationally infeasible. 

\textbf{\ours~resolves the tension between localization and exact unlearning.} Given the above challenges, our goal is to construct local masks in a manner which depends \textbf{only} on the local model and \textbf{not} the merged model. To do this, we propose SIFT (Sign-Fixed Tuning), which is shown in Figure~\ref{fig:sift-figure}. The first step of SIFT is to initialize a uniformly random sign vector $v$ that is shared across all tasks. During local finetuning, we constrain the entries of each task vector $\tau_c$ such that $\tau_c \odot v$ is greater than or equal to 0. We project $\tau_c$ to this constraint set after each gradient step by clipping all entries of $\tau_c$ to 0 where $\tau_c \odot v < 0$. After finetuning, all task vector weights will either be 0 or share the same sign as $v$, which produces the local mask $m_c = \mathbbm{1}\{\tau_c \odot v > 0\}$. Our complete method \ours
(pseudocode in Algorithm~\ref{alg:sift}) 
applies these masks to the merged task vector. In addition to enabling exact unlearning, \ours improves localized model performance over \tall and \emr. This is enabled by how we change the finetuning procedure; across each task $c$, \ours aligns $\tau_c$ to agree on all signs prior to merging, leading to agreement with $\bar\tau$.

\section{Experiment Setup}
In the following sections, we empirically evaluate the performance of \ours. We first consider the accuracy of \ours~for merging at scale in Sec.~\ref{sec:results:accuracy}. We then analyze unlearning costs, measuring the cost of unlearning a single task across methods in Sec.~\ref{sec:results:preunlearn}, and  unlearning multiple tasks in Sec.~\ref{sec:results:postunlearn}. Finally, we evaluate simple alternatives to improve the na\"ive baselines (\ct~and \texttt{FT\;+\;Merge}) in Sec.~\ref{sec:results:storage}.

% \begin{table}[]
% \normalsize
% \centering
% \begin{tabular}{lccccccc}
% \toprule[\heavyrulewidth]
% \textbf{Dataset} & \textbf{Tasks} & \textbf{Labels} & \textbf{Label Type} & \textbf{Total Examples}\\
% \midrule
% TOFU          & 200 & 50256 & tokens & 4000 \\
% Sent140       & 500 & 2 & sentiment & 40504 \\
% Reddit        & 500 & 20 & subreddits & 67484 \\
% StackOverflow & 500 & 10 & tags & 97037 \\
% MATH          & 35  & -  & exact match & 12500 \\
% PubMed        & 100 & 50256 & tokens & 13251 \\
% \bottomrule[\heavyrulewidth]
% \end{tabular}
% % \vspace{-0.2cm}
% \caption{Overview of the datasets used in our experiments. Each dataset is naturally partitioned over a large number of tasks.}
% \label{tab:datasets-small}
% \end{table}

We finetune models ranging from 700M to 1.5B parameters on four common text datasets from the unlearning and federated learning literature, where unlearning requests naturally occur at the task level: TOFU~\citep{maini2024tofu}, Sent140~\citep{go2009twitter}, Reddit~\citep{caldas2018leaf}, and StackOverflow~\citep{mikex86stackoverflow}. Additionally, we consider the more complex MATH~\cite{hendrycks2measuring} and PubMed~\cite{nihPubMed} datasets. 

TOFU is a QA dataset where we evaluate the probability of generating a sequence of tokens corresponding to a given answer. Sent140 is a binary classification task where the goal is to predict the sentiment of a Twitter comment. For StackOverflow and Reddit, we formulate topic classification tasks where the goal is to classify a comment as one of 50 tags or 20 subreddits respectively. MATH requires the model to generate a LaTeX expression containing the final answer, and is often solved by prompting the model to first generate several reasoning outputs. PubMed is a collection of medical papers where we treat each paper as a separate language modeling task. We use PubMed as a more complex language modeling task compared to TOFU, as each task contains roughly 6 times the number of examples compared to that of TOFU.

\begin{table}[]
\normalsize
\centering
\begin{tabular}{lcccc}
\toprule[\heavyrulewidth]
\textbf{Dataset} & \textbf{Tasks} & \textbf{Labels} & \textbf{Label Type}\\
\midrule
TOFU          & 200 & 50256 & tokens \\
Sent140       & 500 & 2 & sentiment \\
Reddit        & 500 & 20 & subreddits \\
StackOverflow & 500 & 10 & tags \\
MATH          & 35  & -  & exact match \\
PubMed        & 100 & 50256 & tokens \\
\bottomrule[\heavyrulewidth]
\end{tabular}
% \vspace{-0.2cm}
\caption{\textbf{Overview of our datasets.} Each dataset is naturally partitioned over a large number of tasks. Sent140, Reddit, and StackOverflow are classification datasets, TOFU and PubMed are language modeling datasets, and MATH is a reasoning dataset.}
\label{tab:datasets-small}
\end{table}

We finetune either GPT2-XL~\citep{radford2019language}, FLAN-T5-Large~\citep{chung2024scaling}, or Llama3.2-1B-Instruct~\citep{grattafiori2024llama}; all are pretrained models ranging from 700M to 1.5B parameters. We provide a few details in Table~\ref{tab:datasets-small} with full dataset \& model details in Appendix~\ref{app:experiments}. This appendix section also contains a discussion on how we selected the specific model for each dataset. In addition, we confirm that \ours~is robust to many settings with an experiment which sweeps over multiple dataset and model pairs.

\section{Pre-unlearning Accuracy}
\label{sec:results:accuracy}
Although model merging may be useful to enable exact unlearning, it is important to ensure that it doesn't come at the cost of reduced accuracy of the merged model. In this section, we evaluate the pre-unlearning accuracy of \ours~and compare to model merging baselines introduced in Table~\ref{tab:compare_methods} (\ft, \tall, \emr). We show that at scale, \ours performs significantly better than \ft and is competitive with the more complex localization methods. Then, we present additional experiments on sparsity including \stitch, showing that \ours outperforms other baselines in terms of model similarity metrics.
% \gs{Can you add an introductory paragraph here to explain what's covered in this section?} \gs{What are the methods we compare to, and why? What are we trying to show?}

\subsection{Merging and localization accuracy}
\begin{figure}
    % \vspace{-0.2cm}
    \centering
        \includegraphics[width=1\linewidth]{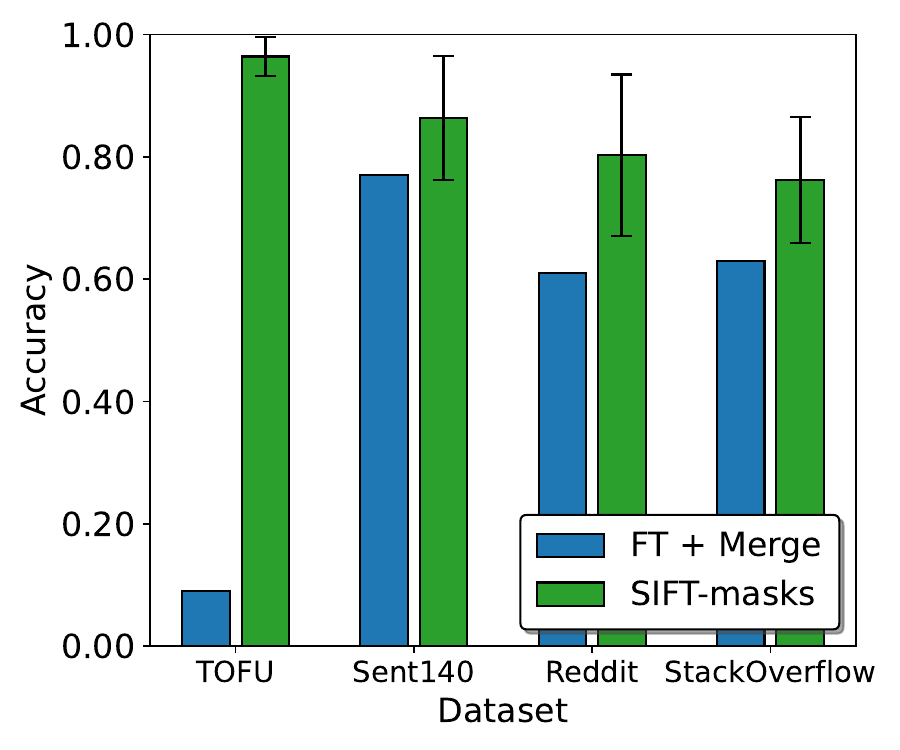}
    % \vspace{-0.2cm}
    \caption{\textbf{\ours outperforms \ft across multiple datasets.} We finetune on 500 tasks for each dataset (200 for TOFU) and evaluate the performance of \ours compared to \ft. Across the 4 datasets, \ours improves performance by 5-80\%. While merging at scale significantly harms \ft in some settings (TOFU), it can retain reasonable performance in other settings (Sent140) since we initialize finetuning with a strong zero-shot model.}
    \label{fig:merging_datasets}
    % \vspace{-0.2cm}
\end{figure}

\begin{figure}[]
    \centering
    % \vspace{-0.4cm}
    \includegraphics[width=0.95\linewidth]{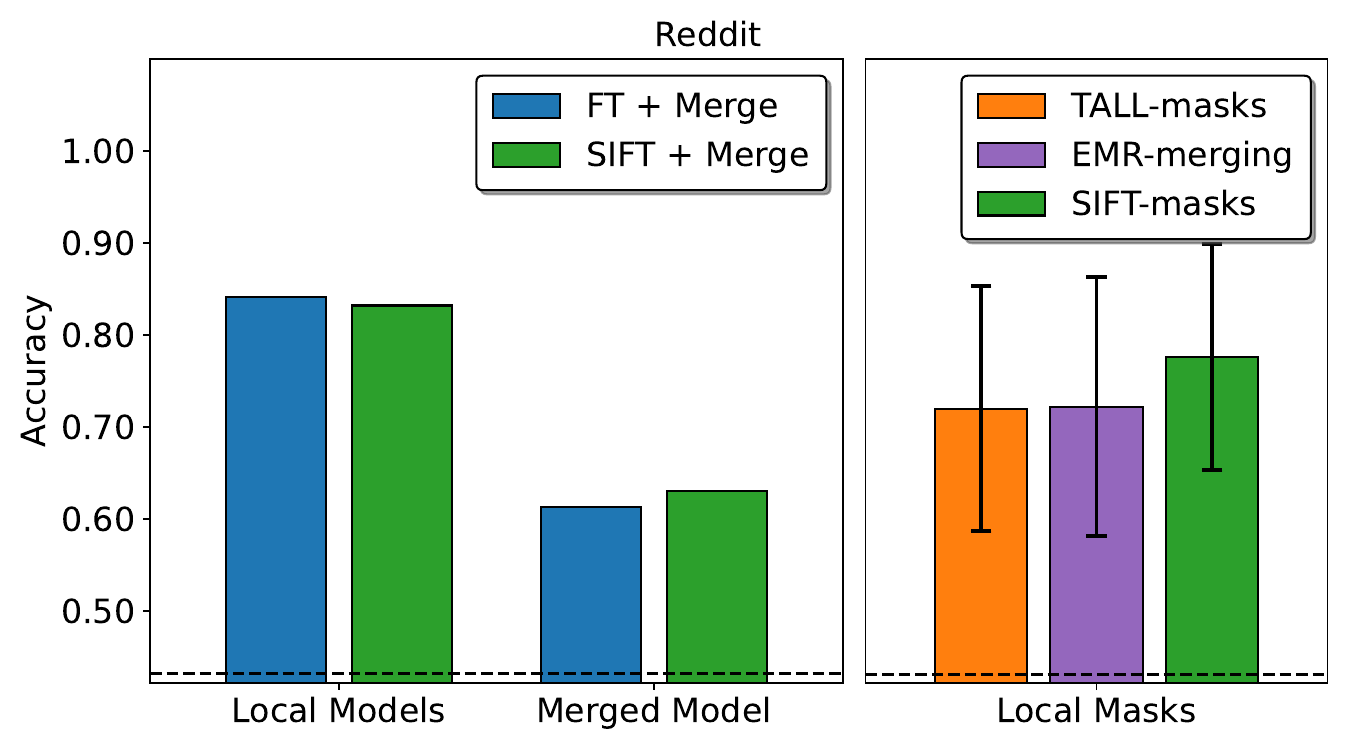}
    \includegraphics[width=0.95\linewidth]{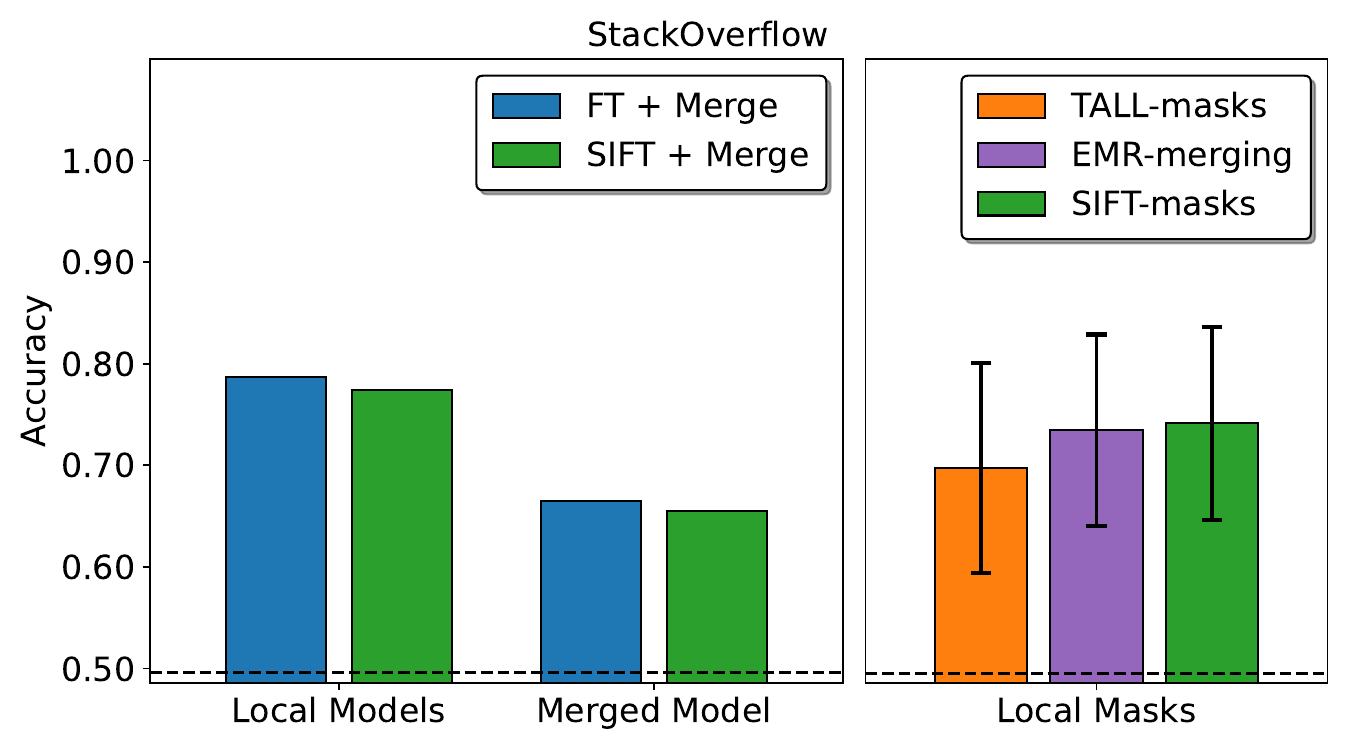}
    % \vspace{-0.2cm}
    \caption{\textbf{\ours~is competitive with \tall and even local models in terms of accuracy.} We finetune 500 tasks on on Reddit and StackOverflow, and evaluate the performance immediately after each step of merging-and-localization: finetuning (local models), merging (merged model), and localization (local models). Surprisingly, we find that \ours outperforms \tall, despite using a relatively simple technique to construct the masks.}
    % \vspace{-0.2cm}
    \label{fig:sift_merge}
\end{figure}

First, we consider na\"ive finetuning and merging (\ft) as a natural baseline for this set of experiments.
% \gs{CAn you remind people what FT+merge is, and why this is the natural baseline to consider? (Eg why don't we compare to central FT as well?} 
In Figure~\ref{fig:merging_intro}, we plot the performance (answer probability) of \ours~and \ft~on TOFU as we vary the number of merged models from 1 to 200. At 200 merged models, \ft~degrades to zero-shot probability, while \ours~remains at 99\% probability. Next, we show that across multiple datasets and varying numbers of merged models, \ours~has significant benefits compared to \ft. In Figure~\ref{fig:merging_datasets}, we show the performance of these two methods at the maximum number of tasks (200 for TOFU, 500 for others). Across the other 3 datasets, \ours~recovers 5-20\% accuracy. Next, we compare \texttt{SIFT + Merge} to \ft at a more detailed level. We show that the sign constraint of SIFT has little impact on finetuning quality in terms of both the local and merged models' performance. In Figure~\ref{fig:sift_merge}, the performance of \texttt{SIFT} is similar to that of \texttt{FT} (\texttt{Local Models}). 
% Although \texttt{SIFT + Merge} eliminates sign conflicts during merging, the large number of tasks we merge (500) still causes significant interference, resulting in similar performance as regular \ft. The benefit of SIFT is only clear once the local masks are applied. 
However, despite starting from finetuned models of similar quality, \ours~outperforms other localization baselines that construct the mask based on the merged weights.
%\tall, a baseline which optimizes the mask to minimize distance between the merged and local models.


We additionally test \ours~on two more complex datasets: MATH and PubMed. We split the MATH dataset~\citep{hendrycks2measuring} into 35 tasks for each topic-difficulty pair and finetune Llama3-1B-Instruct on these tasks. For PubMed, we download 100 articles and finetune GPT2-XL with a language modeling objective on each article (task). Table~\ref{tab:math_pubmed} shows the results, with \ours outperforming both \ft and \tall.

\begin{table}[h]
\centering
\normalsize
\begin{tabular}{lcc}
\toprule
\textbf{Method} & \textbf{MATH (35 tasks)} & \textbf{PubMed (100 tasks)} \\
\midrule
\ft & 18.8 & 37.4 \\
\tall & 21.8 & 51.9 \\
\ours & \textbf{22.6} & \textbf{80.7} \\
\bottomrule
\end{tabular}
\caption{\textbf{Pre-unlearning accuracy on MATH and PubMed.}}
\label{tab:math_pubmed}
\end{table}

\begin{figure}{}
    \vspace{-0.5cm}
    \includegraphics[width=1.0\linewidth]{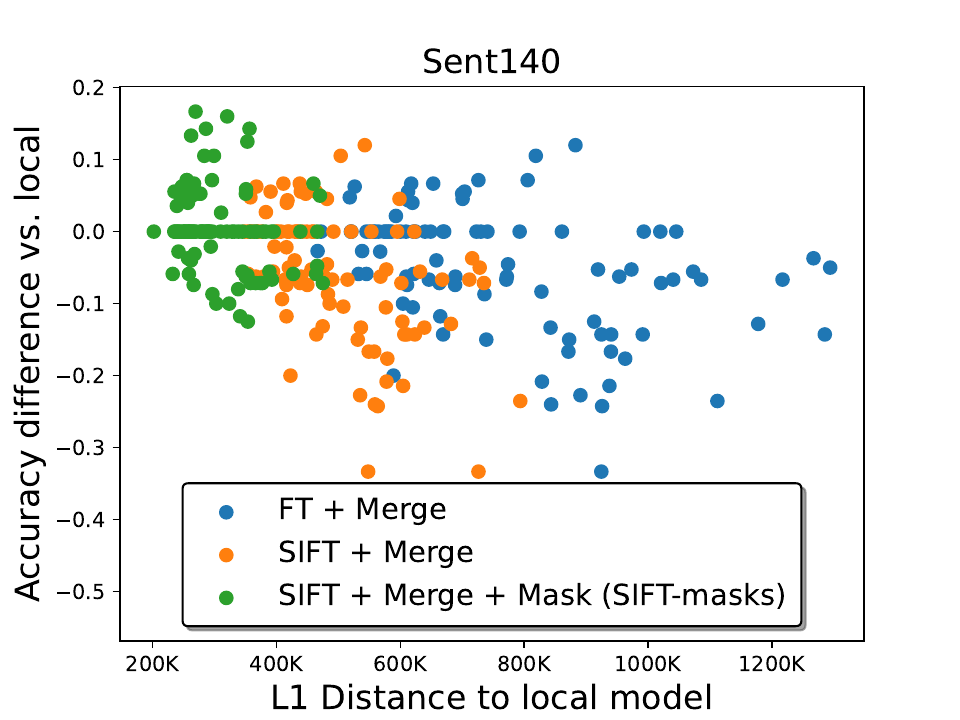}
    % \vspace{-0.6cm}
    \caption{\textbf{SIFT reduces the distance between the merged and local models (blue vs. orange)}. However, this does not directly improve accuracy due to interference from merging a large number of tasks. Instead, accuracy only improves after applying local masks (green).}
    \label{fig:sift_distance}
    \vspace{-0.4cm}
\end{figure}

\subsection{Effectiveness of local mask}
Next, we analyze how the original local model and the final merged-and-localized model differ. First, we consider 100 Sent140 tasks and run three different methods: regular finetuning and merging (\ft), sign-fixed tuning and merging (\texttt{SIFT + Merge}), and the complete \ours~method (\texttt{SIFT + Merge + Mask}). In Figure~\ref{fig:sift_distance}, we plot each task as a single point. On the x-axis, we measure the distance from the merged (and localized) model to the original local model. With \texttt{SIFT + Merge}, we eliminate sign conflicts and obtain a merged model which is relatively closer to the local models compared to \ft. However, this does not significantly improve accuracy compared to \ft. For any given model parameter, the local models are unlikely to independently agree on its randomly global sign. This results in a dense merged model containing many weight entries where the local weight has been set to zero via the global sign vector, but the corresponding weight in the merged model has the opposite sign from the gradient of the local model. In other words, the global parameter sign points in a direction which is harmful for that task. \ours~adds task-specific masks which removes these weights and is key to improving accuracy.

\subsection{Naive efficient localization fails}
Since prior works in model merging study how sparsity affects merging, a natural question is whether these methods can be augmented with localization and potentially outperform existing merging-and-localization methods. A representative technique called \stitch \cite{wang2024localizing} proposes merging sparse local models. 
% The justification for this method is that only a small fraction of parameters are responsible for task-specific knowledge, and thus sparsifying the task vectors can reduce interference in the merged model. 
Critically, this approach satisfies a key requirement of merging for exact unlearning: the masks are determined entirely locally. Therefore, to improve \stitch (merge-only), we store the sparse masks generated during the initial finetuning and use them to localize the merged model on a task-specfic basis. We call this resulting method \stitch~(merge-and-localize).

\begin{table}[]
\centering
\normalsize
\begin{tabular}{lccc}
\toprule
\multirow{2}{*}{\textbf{Method}} & \textbf{L1} & \textbf{Sign} & \multirow{2}{*}{\textbf{Probability}} \\
 & \textbf{Distance} & \textbf{Conflicts} & \\
\midrule
\ft & 606K & 587M & 0.09 \\
\multirow{2}{*}{\shortstack[l]{\footnotesize\stitch\\ (merge-only)\normalsize}} & \multirow{2}{*}{561K} & \multirow{2}{*}{95M} & \multirow{2}{*}{0.05} \\
\normalsize &&&\\
\multirow{2}{*}{\shortstack[l]{\footnotesize\stitch\\ (merge-and-localize)\normalsize}}  & \multirow{2}{*}{203K} & \multirow{2}{*}{95M} & \multirow{2}{*}{0.24} \\
 &&& \\
\tall & 549K & 0 & 0.31 \\
\ours & \textbf{223K} & \textbf{0} & \textbf{0.94} \\
\midrule
\end{tabular}
\caption{\textbf{We compare the final served model to the original local model for each task.} \ft and \stitch (merge-only) only produce a single model and perform poorly, while localization slightly improves \stitch.
% and the merged model conflicts with close to 50\% of parameters (600M out of 1.2B). 
% Localization improves \stitch, but it still lags behind methods originally designed with both merging and localization in mind.
}
\label{tab:sign_conflicts}
\end{table}
% \vspace{-0.4cm}

\begin{table}[]
\centering
\normalsize
\begin{tabular}{lccc}
\midrule
\textbf{Method} & \textbf{Density} & \textbf{FT+Merge} & \textbf{+Mask} \\
\midrule
\stitch \\
-- Dataless & 0.05 & 0.01 & 0.20 \\
-- Dataless & 0.20 & 0.05 & 0.24 \\
-- $\lambda$=100 & 0.40 & 0.09 & 0.13 \\
\ours & \textbf{0.57} & \textbf{0.09} & \textbf{0.94} \\
\midrule
\end{tabular}
\caption{\textbf{We vary how local masks are determined in \stitch.} If data is available, the model can directly be optimized using SGD with sparsity regularization ($\lambda=100$). In dataless settings, the mask retains the TopK\% of local weights.}
\label{tab:localize_stitch}
\end{table}
% \vspace{-0.2cm}

We compare \stitch to other baselines in Table~\ref{tab:sign_conflicts}, which measures response probability on TOFU. First, incorporating localization improves \stitch by almost 20\% probability (0.05 to 0.24), and reducing sign conflicts and L1 distance over \ft. 
% Both ``merge-only'' and ``merge-and-localize'' significantly reduce sign conflicts and L1 distance respectively over \ft. 
However, despite the overall effectiveness of \stitch and \tall at reducing these metrics, their task performance is significantly lower than \ours (a gap of 0.31 vs. 0.94 probability). This is because the sign conflict metric can at times be misleading; since zero vs. non-zero is not counted as a sign conflict, sparser models (even randomly sparse) naturally have less sign conflicts. Finally, in Table~\ref{tab:localize_stitch}, we tested a few hyperparameter configurations of \stitch in order to determine the best configuration (dataless, $\lambda=0.20$) to present in Table~\ref{tab:sign_conflicts}. Overall, these experiments show that we must carefully consider how sparsity is applied to merging-and-localization, as na\"ively masking only local models (merge-only), only the merged model (\tall), or even tailoring existing methods for localization may be insufficient.

\begin{figure*}[]
    \centering
    \vspace{-0.4cm}
    \includegraphics[width=1.0\linewidth]{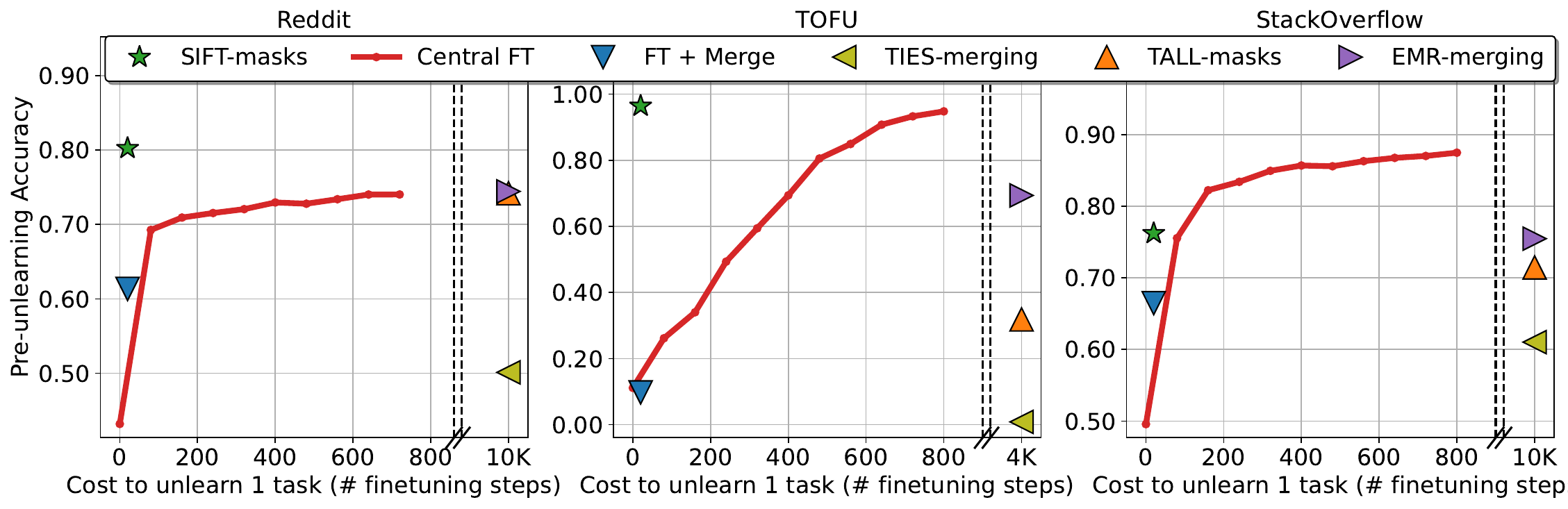}
    % \vspace{-0.2cm}
    \caption{\textbf{We compare the cost to unlearn 1 task from \ours~versus \ct, \ft, and \tall.} Depending on the setting, \ours~can outperform \ct~due to task heterogeneity (e.g. conflicting examples or distinct distributions). Across all settings, \ours~improves on the tradeoff of efficiency and accuracy compared to varying the number of finetuning steps used for \ct. Compared to merging-and-localization baselines, \ours is not only competitive in terms of accuracy, but also reduces unlearning costs by multiple orders of magnitude.}
    \label{fig:finetune-vary-central}
    \vspace{-0.2cm}
\end{figure*}

\begin{mybox}{Takeaway}
By applying a global sign constraint during finetuning, \ours improves the accuracy of merging-and-localization methods which otherwise degrade at scale.
\end{mybox}

\section{Unlearning Costs}
\label{sec:results}

Understanding unlearning costs is critical for deployment of ML systems that must comply with data deletion requests. In this section, we evaluate the trade-off between compute efficiency and model accuracy. We examine this from two perspectives: the immediate cost of unlearning individual tasks, and the accumulated cost of handling multiple sequential deletion requests. Our analysis reveals that while merging approaches require higher upfront costs, they provide substantial efficiency during unlearning---a tradeoff that becomes increasingly favorable as the number of total tasks grows.
% \gs{Can you add an intro paragraph here explaining what will be covered at a high level, and reminding them why unlearning costs are important?}

\subsection{Finetuning cost}
\label{sec:results:preunlearn}

In this section, we compare the unlearning cost of each method. In terms of the number of finetuning steps required for exact unlearning, \ours~is more efficient than all other methods: \ct, \ft, \ties~\citep{yadav2024ties}, \tall~\citep{wang2024localizing}, and \emr~\citep{huang2024emr}. We finetune for up to 800 steps on \ct~and fix 20 steps for all other approaches. For $T$ tasks, this is a total of $20T$ finetuning steps (4,000 for TOFU, 10,000 otherwise). Although this initial cost is high, it allows \ft~and~\ours~to efficiently unlearn a task via finetuning on and unmerging that task. However, as previously mentioned, this same technique does not result in exact unlearning for the other methods.

\begin{figure}
    \vspace{-0.5cm}
    \includegraphics[trim={0 1cm 0 0.7cm},clip,width=1.0\linewidth]{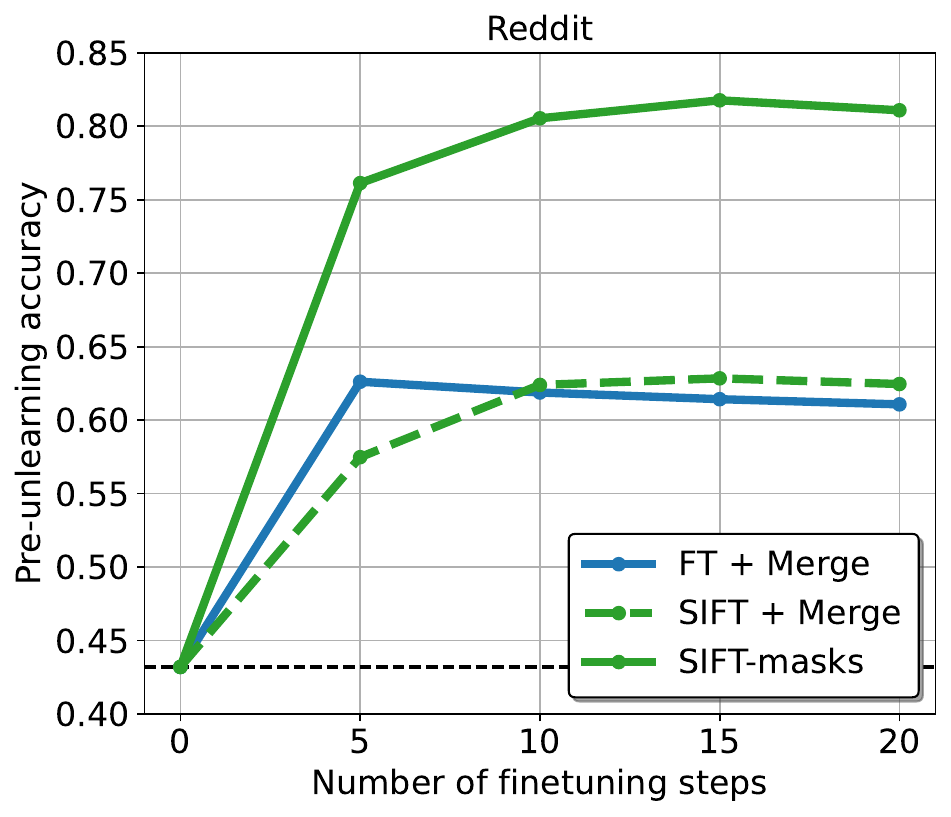}
    % \includegraphics[width=1.0\linewidth]{figures/5b-finetune-vary-merge-reddit.pdf}
    % \vspace{-0.1cm}
    \caption{\textbf{We vary the number of steps used for local finetuning.} While we use 20 steps for local finetuning on all datasets and tasks, this amount can be carefully reduced to make merging approaches even more efficient.}
    \vspace{-0.3cm}
    \label{fig:finetune-vary-merge}
\end{figure}

\begin{figure*}[]
    \vspace{-0.4cm}
    \centering
    \includegraphics[width=0.5\linewidth]{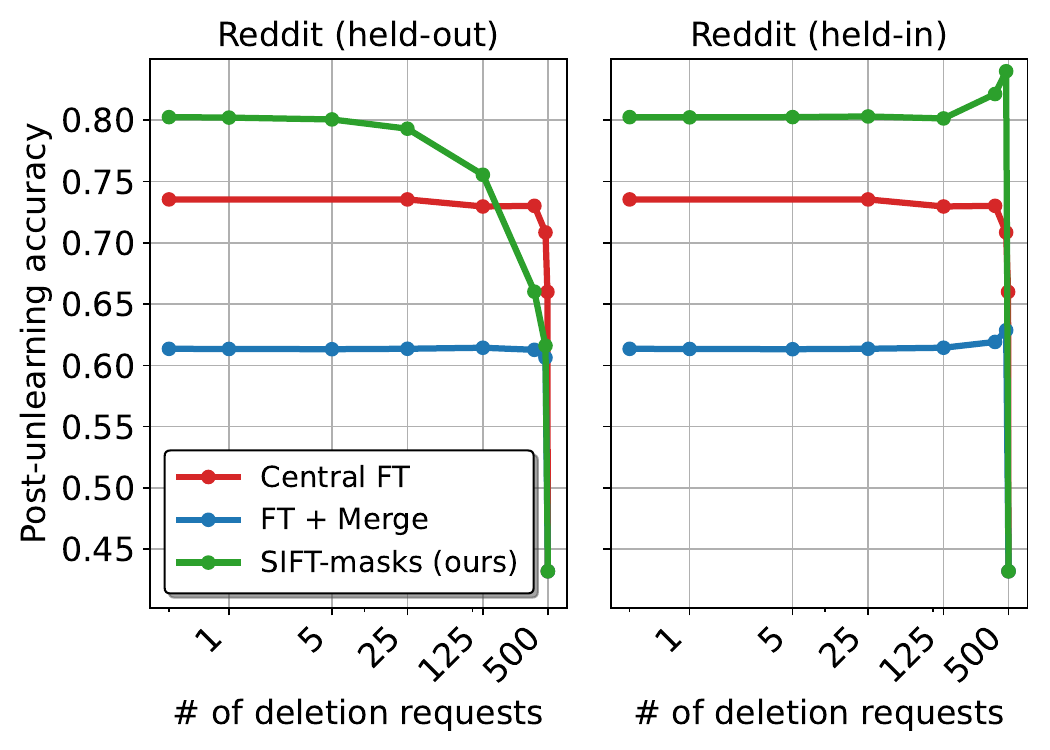}~    \includegraphics[width=0.5\linewidth]{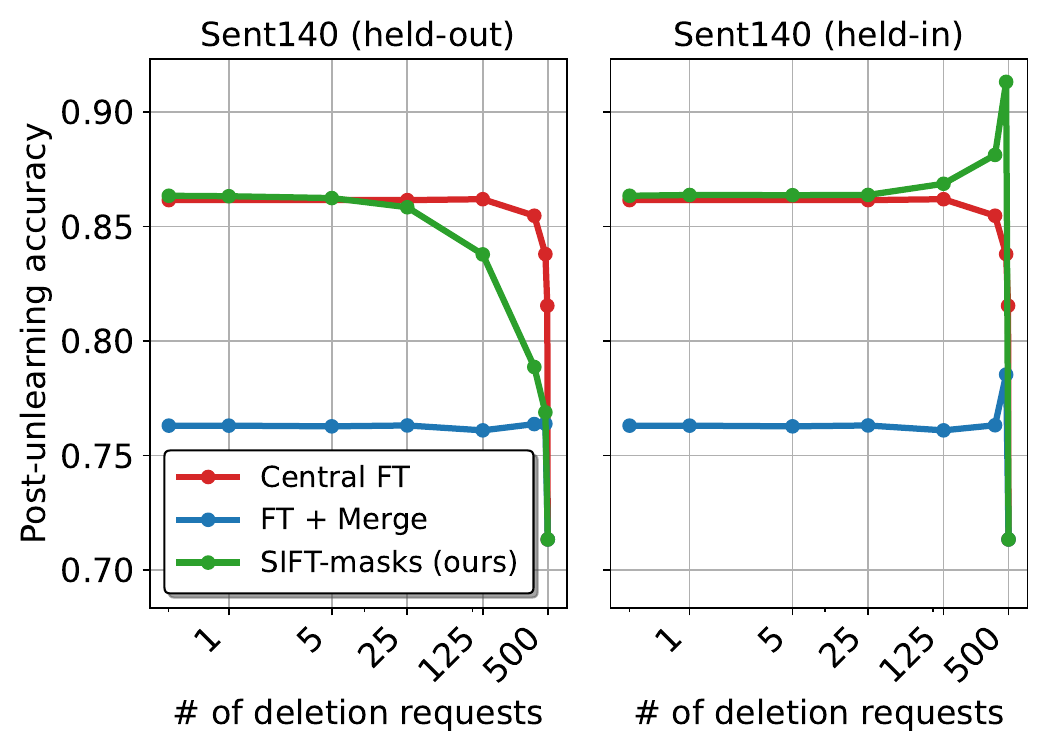}
    \caption{We evaluate two post-unlearning metrics: \emph{held-out} performance on all clients (including unlearned clients) and \emph{held-in} performance on the retain clients. For unlearned clients, \ours~applies the merged model without any mask. Held-out performance decreases as more tasks are unlearned, due to the poor performance of the merged model compared to \ours. Held-in performance increases as more tasks are unlearned because of less interference in the merged model. }
    \label{fig:forget-utility}
    \vspace{-0.2cm}
\end{figure*}

In Figure~\ref{fig:finetune-vary-central}, the setting can affect how quickly \ct~converges and its accuracy relative to \ours. We conjecture that this is due to \emph{data heterogeneity}: the degree to which the data for one task is helpful for another task. On Reddit, tasks have potentially \emph{conflicting} data which a single model cannot handle; for example, similar generic comments can be posted across different subreddits. On TOFU, each task (author) contains highly \emph{distinct} data; training on one task has little to no affect on the other tasks. Since TOFU measures held-in training performance, \ct~is guaranteed to reach 100\%, but converges slowly due to the structure of the data. Finally, data across StackOverflow tasks is \emph{similar}; 
labels in StackOverflow (e.g. programming languages) depend less on the task context (user) and more on global features (e.g. language keywords) of the comment. As a result, \ours~improves efficiency on StackOverflow, but cannot reach the performance of centralized training. Overall, these results show that \emph{merging is most helpful when data is heterogeneous} i.e. performing well on a given task requires knowledge of the context or having finetuned on its training data. Additionally, we show that existing localization methods are an extremely poor choice for exact unlearning. Due to the nature of these localization methods, unlearning a single task requires re-running the entire merging and localization method on the remaining 499 tasks, resulting in $499\times20=9980$ finetuning steps and costing several times more than naive \ct. Finally, \ties~(Section~\ref{sec:relwork}) performs poorly because it does not use localization, similar to \texttt{SIFT + Merge} in Figure~\ref{fig:sift_distance}. In Appendix~\ref{app:experiments}, we provide finer-grained experiments comparing \ties~to \ft.

For all datasets, we found that tuning the number of steps individually for each task was generally not helpful and that 20 finetuning steps worked well as a fixed quantitiy across all tasks. On 100 Reddit tasks, we run an additional ablation on the amount of finetuning steps. In Figure~\ref{fig:finetune-vary-merge}, we show that it is possible to use as few as 10 finetuning steps per task and achieve similar accuracy after merging and applying~\ours. We also note that the sign constraint slightly slows down training, as shown by the performance of~\ft~versus~\texttt{SIFT + Merge} at $x=5$. However, 
% as discussed, \ours~does not directly apply the merged model and instead combines it with a mask. Therefore, 
for the same number of finetuning steps,~\ours~always outperforms~\ft.

\begin{figure}
    \vspace{-0.3cm}
    \includegraphics[width=1.0\linewidth]{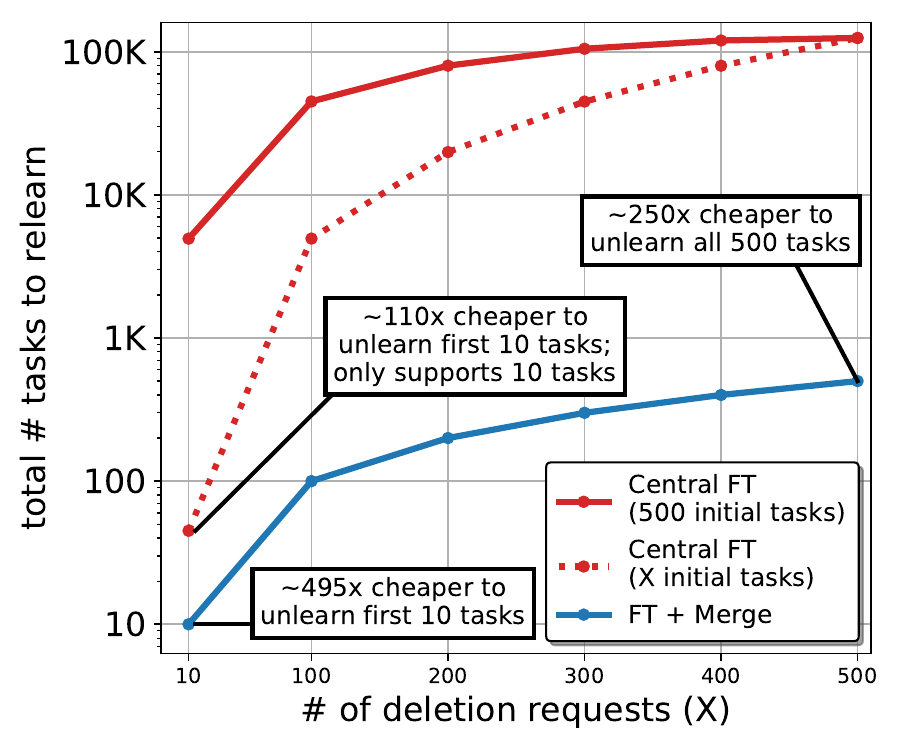}
    \caption{Training on fewer initial tasks ($X < 500$) can reduce the unlearning costs of \ct, but limits the maximum number of deletion requests the system can support. Furthermore, the total cost of unlearning is still $X/2$ to $X$ times larger than that of merging approaches. Y-axis is on a log scale.}
    \label{fig:forget-relearn-cost}
    \vspace{-0.4cm}
\end{figure}

\begin{figure*}
    \vspace{-0.4cm}
    \centering
    \includegraphics[width=0.9\linewidth]{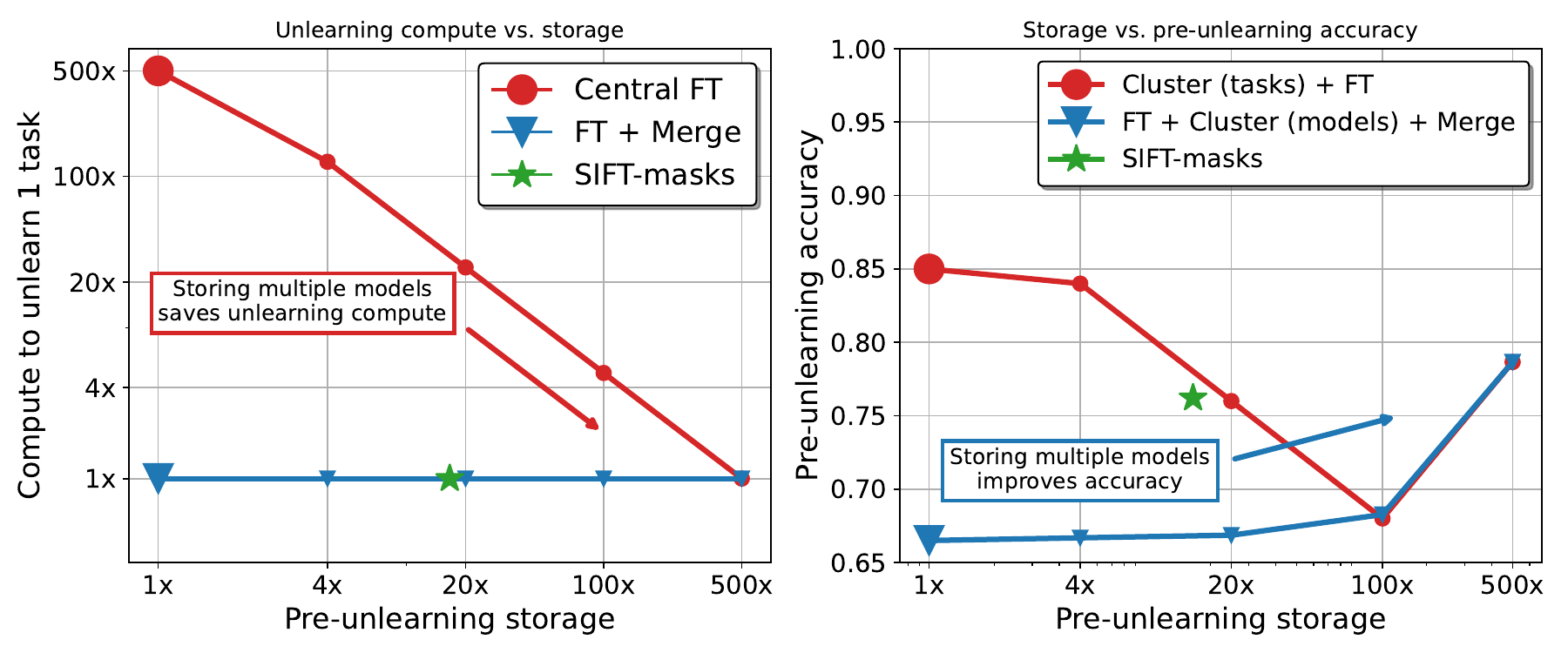}
    \vspace{-0.2cm}
    \caption{\textbf{Clustering allows na\"ive baselines to trade off between accuracy and resource constraints.} By paying extra storage cost to maintain multiple models, we can reduce the unlearning cost of \ct~(left) or improve the accuracy of \ft~(right). However, storing additional models with these two baselines is unable to match the compute or accuracy benefits of \ours.}
    \label{fig:storage}
    \vspace{-0.2cm}
\end{figure*}

\subsection{Handling multiple unlearning requests}
\label{sec:results:postunlearn}
Next, we evaluate the performance of various methods as multiple tasks are unlearned, ranging from 1 to all 500 tasks. In Figure~\ref{fig:forget-utility}, we show two different evaluation metrics: the first is a \emph{held-out} evaluation that includes all 500 tasks in the dataset. To evaluate a task which has already been unlearned, \ours~applies the merged model without any mask. Therefore, as tasks are unlearned, the global performance of \ours~\textbf{decreases} as clients are unlearned and eventually matches the performance of \ft. The other metric is a \emph{held-in} evaluation that only includes the retained tasks. Unlike the held-out setting, each task always has a local mask \ours can apply. In this setting, performance \textbf{increases} as clients are unlearned. This is because of less interference during merging (Fig.~\ref{fig:merging_intro}), allowing \ours~to obtain better masked models. Once all clients are unlearned, performance for all methods drops to zeroshot accuracy.

When deletion requests arrive iteratively, satisfying these requests one-by-one is difficult for \ct~and leads to a large total cost. In Figure~\ref{fig:forget-relearn-cost}, we show how the cost of relearning is accumulated across up to 500 unlearning requests. We measure costs in terms of the number of tasks that require finetuning for unlearning and relearning. This metric assumes that it is necessary to train on a task's data in order to perform well on its evaluation set. While this only true in certain settings (like on the TOFU dataset), it clearly highlights the limitations of \ct. For example, if the unlearning system starts with $500$ tasks, unlearning the first task with model merging only requires relearning that single task and unmerging its model. However, unlearning with \ct~requires learning all 499 tasks in the retain set. As more tasks are unlearned beyond this first task, this ratio becomes smaller due to the shrinking size of the retain set and reaches $250\times$ once all $500$ tasks are unlearned. However, in absolute terms, this difference is very large: \ft~only needs to relearn $499$ tasks in total, while \ct~needs to relearn $\sum_{i=1}^{499} (500-i) \approx 125,000$ tasks. While reducing the amount of initial tasks can significantly reduce the relearning costs, this limits the number of tasks (deletion requests) the system can support. Even if we reduce the number of initial tasks to $X$, the cost of unlearning is still at least $(X-1)/2$ times more expensive than merging.

\begin{mybox}{Takeaway}
\ours improves on resource tradeoffs of na\"ive baselines, providing strong performance as tasks are unlearned, cheaper unlearning costs, and increased capacity for deletion requests.
\end{mybox}

\section{Storage Costs} 
\label{sec:results:storage}

Storage overhead is a key consideration when deploying unlearning systems at scale, since these costs can become prohibitive as more tasks are added to the finetuning dataset. In this section, we analyze the storage costs of different methods and evaluate simple strategies to reduce these costs. We show that clustering allows merging methods to trade off between accuracy and unlearning costs, while LoRA (low-rank adaptation)~\citep{hu2022lora} allows scaling to an even larger number of tasks while keeping storage costs reasonable.

\subsection{Trading compute for storage}
We compare storage costs across methods and discuss the tradeoff between storage, unlearning compute, and pre-unlearning accuracy. A naive way to improve the baselines of \ct~and \ft~is to allow storing multiple models. Instead of a single \ct~model, we randomly cluster the tasks into equal-sized clusters, then train a ``local'' model for each cluster, and then simply serve each local model for any task in its cluster. Similarly, instead of a single \ft~model, we randomly cluster the (task-level) local models and merge the models in each cluster. For a given number of clusters, both of these modified methods use the same amount of storage, but vary in their compute cost and performance. The benefit of clustering in \ct~is to reduce unlearning compute, since fewer tasks have to be relearned in a given cluster. On the other hand, the benefit of clustering in \ft~is to improve accuracy, since we merge less models at once and reduce merging interference. In both cases, we use random clustering because data-driven clustering cannot be trivially unlearned.

In Figure~\ref{fig:storage}, we first show in the left plot that localization approaches (\ours~and \tall) cost additional storage compared to single-model methods \ct~and \ft. For 500 tasks, this storage cost is $500/32 \approx 16$ times that of a single model. Next, we show that clustering the tasks has limited benefits for either of the two baselines. First, we focus on how forming clusters of tasks affects finetuning within each cluster (\ct). On the left plot, we show that learning a \ct~model for each cluster reduces unlearning costs (y-axis), but these costs remain several times greater than the cost of \ours~or~\ft. Furthermore, as shown in the right plot, clustering in this way harms the performance of \ct; . Next, we focus on how forming clusters of local models affects merging within each cluster (\ft). In the left plot, we show that clustering only increases storage costs while unlearning costs stay constant. This is because the procedure to unlearn a given task is exactly the same as in regular \ft (i.e. only 1 cluster), except that merging/unmerging is only done within a given cluster of models. On the right plot, clustering improves the accuracy of the cluster-level merged models, but accuracy remains worse than \ours until we have 500$\times$ clusters, i.e. each task is its own cluster.

\subsection{Varying Base Model}
% \gs{ADd more text to this section -- why are we looking at these experiments? What is the takeaway?}
Next, evaluate our approach across different model sizes. In Table~\ref{tab:model_size}, we compare FLAN-T5-Small (70M params) to FLAN-T5-Large (700M params) across Reddit and StackOverflow datasets. As expected, larger models consistently improve performance across all methods. Importantly, our method increases its advantage over baselines as models grow larger, suggesting that the benefits of sign-constrained merging and localization are robust to model scale. This indicates that our approach should continue to be effective as models grow larger.
 
\subsection{LoRA}
% \gs{Add more text here -- why are we looking at these experiments/whats the take-away?}
LoRA (low-rank adaption)~\cite{hu2022lora} is a parameter-efficient finetuning method which reduces per-task storage from full model parameters to a lightweight ``adapter'' module. In Table~\ref{tab:lora}, we compare LoRA (with rank r=16) to full finetuning on Reddit and StackOverflow. While LoRA slightly reduces performance compared to full finetuning, our method consistently outperforms baselines in both settings.

\subsection{Scaling to 2500 tasks}
To demonstrate the practical scalability of our approach, we evaluate performance when scaling to 2500 tasks---5× more than our standard experiments. This stress test is crucial for understanding whether our method maintains its advantages at the scale required for real-world deployment, where systems may need to handle thousands of users or data sources. In Table~\ref{tab:lora_scale}, we show results for FLAN-T5-Large on Reddit with 2500 tasks using LoRA adapters. At this scale, our method achieves 82.6\% accuracy while requiring only 3.01 GB of storage, a $\sim 30\times$ improvement over storing all local models. 

\begin{table}[]
\centering
\normalsize
\begin{tabular}{lcc}
\toprule
\textbf{Method} & \textbf{flan-t5-small} & \textbf{flan-t5-large} \\
\midrule
\multicolumn{3}{c}{\textbf{Reddit}} \\
\midrule
\ft & 0.22 & 0.59 \\
\tall & \textbf{0.29} & 0.74 \\
\ours & 0.29 & \textbf{0.78} \\
\midrule
\multicolumn{3}{c}{\textbf{StackOverflow}} \\
\midrule
\ft & 0.40 & 0.66 \\
\tall & 0.40 & 0.70 \\
\ours & \textbf{0.44} & \textbf{0.72} \\
\bottomrule
\end{tabular}
\caption{\textbf{We test our approach on two sizes of the FLAN-T5 architecture.} Unfortunately, larger pretrained models are critical for strong performance.}
\vspace{-0.2cm}
\label{tab:model_size}
\end{table}

\begin{table}[]
\centering
\normalsize
\begin{tabular}{lcc}
\midrule
\textbf{Method} & \textbf{LoRA} $(r=16)$ & \textbf{Full FT} \\
\midrule
\multicolumn{3}{c}{\textbf{Reddit}} \\
\midrule
\ft & 0.57 & 0.61 \\
\tall & 0.73 & 0.74 \\
\ours & \textbf{0.76} & \textbf{0.80} \\
\midrule
\multicolumn{3}{c}{\textbf{StackOverflow}} \\
\midrule
\ft & 0.66 & 0.66 \\
\tall & 0.69 & 0.71 \\
\ours & \textbf{0.70} & \textbf{0.76} \\
\midrule
\end{tabular}
\caption{\textbf{We combine LoRA with merging approaches.} While the gap between merging methods is larger under Full FT, \ours retains better performance than the baselines.}
\label{tab:lora}
\end{table}

\begin{table}[]
\centering
\normalsize
\begin{tabular}{lcc}
\midrule
\textbf{Method} & \textbf{Accuracy} & \textbf{Storage Cost} \\
\midrule
Local LoRAs & 0.86 & 95 GB \\
\midrule
\ft & 0.59 & \textbf{38.0 MB} \\
\ct & 0.74 & \textbf{38.0 MB} \\
\emr & 0.72 & 3.01 GB \\
\tall & 0.73 &  3.01 GB \\
\ours & \textbf{0.82} & 3.01 GB \\
\midrule
\end{tabular}
\caption{\textbf{LoRA enables us to scale to 2500 tasks.} Without LoRA, the storage costs would be $\sim70\times$ larger, meaning that storing all local models would cost $\sim$7TB.}
\label{tab:lora_scale}
\end{table}

\begin{mybox}{Takeaway}
Merging methods such as \ours can be combined with orthogonal methods to reduce storage, such as clustering and parameter-efficient finetuning. In these settings, \ours provides better trade-offs between accuracy, storage, and unlearning cost.
\end{mybox}

\section{Conclusion and Future Work}
In this work, we propose using model merging and localization for exact unlearning at scale. While merging is a natural framework for exact unlearning, using simple merging strategies such as averaging results in poor accuracy when a large number of models are merged, making additional techniques such as localization necessary. Unfortunately, these additional techniques can in turn make unlearning expensive. To address these limitations, we propose \ours, a method that uses sign-fixed finetuning to construct local masks without depending on any cross-task information. As a result, \ours~improves the quality of merging while retaining its unlearning efficiency at scale. 

Overall, our work makes an important first step in identifying both the strengths and limitations of using model merging to produce models that are amenable to exact unlearning. Our results suggest that to make model merging even more effective for unlearning, future work should focus on (1) ways to improve the quality of the merged model and (2) localization methods which are amenable to efficient unlearning. Finally, it is important to further explore the limitations of merging in general for efficient ML systems; for instance, prior work suggests that merging requires strong pretrained models. Understanding such general conditions can enable future methods with even more favorable merging quality.

\pagebreak

\textbf{Acknowledgements.} The authors would like to thank Pratiksha Thaker and Saurabh Garg for insightful discussions on this work. This work was supported in part by the National Science Foundation grants IIS2145670 and CCF2107024, and funding from Amazon, Apple, Google, Intel, Meta, and the CyLab Security and Privacy Institute. Any opinions, findings and conclusions or recommendations expressed in this material are those of the author(s) and do not necessarily reflect the views of any of these funding agencies. AS is thankful for the generous support of JP Morgan AI PhD Fellowship.

% \section*{Acknowledgment}

\printbibliography
\clearpage
\appendices
% \section{Algorithm Pseudocode}
% \label{app:pseudocode}
\section{Experiment Details}
\label{appendix:experiment_details}
\textbf{Algorithm details.} One key detail of \ours~is that finetuning must be fully deterministic. In order to guarantee that a model is properly unlearned, we must obtain the original weights that were used during merging and subtract them from the multi-task vector. 

% \input{sections/pseudocode}
% \label{alg:sift}

% \section{Additional Details for Experiments}
\label{app:setup}
\textbf{Reproducibility Statement} In this section, we include details on our experimental setup. Each method uses a fixed set of hyperparameters in all the experiments it appears in; we only use different hyperparameters depending on the dataset and model. We provide all relevant information on algorithm details and hyperparameter configurations in order to reproduce Algorithm~\ref{alg:sift}. 
% Upon acceptance of this work, we will publically share our source code and data preprocessing setup.
Our source code can be found at \url{http://github.com/imkevinkuo/sift-masks}.

\textbf{Dataset overview.} In Table~\ref{tab:datasets-full}, we provide more complete details on the datasets and models used in our experiments. GPT2-XL has 1.5B parameters, Llama3.2-1B-Instruct has 1.2B parameters, and FLAN-T5-Large has 700M parameters. We run full finetuning on all hidden layers of the model i.e. freeze the embedding / language modeling head.

\textbf{Data access and preprocessing.} Next, we briefly explain the setup for each dataset.

\textbf{TOFU.} The TOFU dataset is readily accessible from Huggingface:~\url{https://github.com/locuslab/tofu}. We do not run any preprocessing or filtering on this dataset.

\textbf{Sent140.} The download script for Sentiment140 can be found on Huggingface at \url{https://huggingface.co/datasets/stanfordnlp/sentiment140}. We do not run any preprocessing or filtering on this dataset.

\textbf{Reddit.} We use a dump of Reddit comments from December 2017. A public version of this data can be found as in the LEAF benchmark dataset at \url{https://github.com/TalwalkarLab/leaf/tree/master/data/reddit}. For our experiments, we obtained the original dump from \url{academictorrents.net} and will share our preprocessed version once the paper is made public.
\begin{enumerate}
    \item For our preprocessing, we identified 20 popular subreddits, excluding a few subreddits where there was significant overlap of content (e.g. ``r/politics'' and ``r/news'').
    \item Our final list of subreddits is as follows: \texttt{['politics', 'nfl', 'nba', 'soccer', 'Bitcoin', 'StarWars', 'DestinyTheGame', 'movies', 'leagueoflegends', 'fantasyfootball', 'hockey', 'teenagers', 'RocketLeagueExchange', 'MMA', 'FireEmblemHeroes', 'NintendoSwitch', 'Overwatch', 'relationships', 'pathofexile', 'anime']}.
    \item We then sampled the top 500 users with the most posts across all these subreddits while excluding accounts containing keyphrases indicating that they are bots. 
    \item The list of bot keywords we used was: \texttt{["auto", "mod", "bot", "ImagesOfNetWork", "MTGCardFetcher", "tippr", "DreamProcessor", "Mentioned\_Videos", "keepdankmemesdank", "User\_Simulator", "AreYouDeaf", "ThisCatMightCheerYou", "TotesMessenger", "transcribersofreddit", "xvicsagex", "notifier", "Roboragi", "robot"]}.
\end{enumerate} 

\textbf{StackOverflow.} The data for StackOverflow can be found at Huggingface at \url{https://huggingface.co/datasets/mikex86/stackoverflow-posts}. We first identified the following top 10 tags: \texttt{['javascript', 'python', 'java', 'c\#', 'php', 'android', 'html', 'jquery', 'c++', 'css']}. We then subsampled the data to only include comments which were labeled with only one of these 10 tags. From this subset, we then sampled the 500 users with the most comments. 

\textbf{Prompting and finetuning setup.} For each dataset, we include a prompt which allows the model to achieve non-trivial zeroshot accuracy. Each model is then trained using a causal language modeling objective. That is, instead of randomly initializing and finetuning a classifier head, we keep the original language modeling head and train the model to output the text corresponding to a label i.e.``positive'' or ``negative'' for Sent140, or one of the subreddit or tags listed above for Reddit or StackOverflow respectively.

\textbf{TOFU.} Each example in TOFU is a question-answer pair $\texttt{(q,a)}$ which we transform using a simple prompt: ``\texttt{Question:\{q\}}\textbackslash n\texttt{Answer:\{a\}}''. We then finetune the model to predict the tokens corresponding to the answer sequence \texttt{a}.

\textbf{Sent140.} Each example is a comment with a binary label indicating whether the comment is positive or negative sentiment. We format this using the following Llama3 prompt: 
\begin{lstlisting}
<|begin_of_text|><|start_header_id|>system<|end_header_id|>
You are an AI assistant designed for sentiment analysis.
<|eot_id|><|start_header_id|>user<|end_header_id|>
What is the sentiment of this comment? Respond with a single word.
Do not start with 'The'. Choices:
- Negative
- Positive
Comment: {comment}
<|eot_id|><|start_header_id|>assistant<|end_header_id|>
{label}
\end{lstlisting}

\textbf{Reddit and StackOverflow.} Like for Sent140, we provide a prompt that includes the list of 20 subreddits or 10 tags we filtered during preprocessing. 
\begin{verbatim}
Context:{comment}
Question:Choose the most relevant topic.
OPTIONS:
{option_list}
Answer: {label}'
\end{verbatim}

\textbf{Hyperparameters.} Next, we list out the hyperparameters that we tuned for our method and other baselines in the experiments. Generally, we used the same values for all shared hyperparameters across methods (i.e. learning rate, batch size, finetuning steps).

\textbf{Finetuning.} As stated in Section~\ref{sec:results:preunlearn}, we used 20 finetuning steps for each individual task and up to 800 finetuning steps for \ct. For TOFU, we finetune with a batch size of 20 (each task contains exactly 20 examples). For the other three datasets, we finetune with a batch size of 128; if a task has less than 128 examples, we use full-batch gradient descent. For TOFU, we use a learning rate of 1e-4. For Reddit and StackOverflow, we use a learning rate of 1e-5, and for Sent140, we use a learning rate of 1e-7.

\textbf{Merging.} In our work we use a simple unweighted average of task vectors, as shown in the ``Localization'' step of Algorithm~\ref{alg:sift}. Although prior works in model merging propose rescaling the average or using a weighted average of the task vectors, we found that these additional techniques were costly to tune and did not result in significant improvements.

\textbf{Localization.} For \tall, we tuned the threshold hyperparameter $\lambda$ based on the density of the mask rather than the threshold value, in order to account for varying magnitudes of merged weights across datasets. We performed a grid search over density values \texttt{[0.1, 0.3, 0.5, 0.7, 0.9]}. For a fair comparison to our method, we consider rescaling the masked model by a hyperparameter $\alpha$ which is also tuned via grid search over values \texttt{[0.8, 1, 1.2, 1.4, 1.6]}. We also test this grid of rescaling values for \emr.

\begin{table*}[]
\centering
\normalsize
\begin{tabular}{lcccccccc}
\toprule[\heavyrulewidth]
& & & \multicolumn{2}{c}{Labels (per task)} & \multicolumn{3}{c}{Examples (per task)} \\
Dataset & Model & & Max & Total & Min & Max & Total \\
\midrule
TOFU          & GPT2-XL & 200 & - & 50256 & 20 & 20 & 4000 \\
Sent140       & Llama3.2-1B-Instruct & 500 & 2 & 2 & 52 & 440 & 40504 \\
Reddit        & FLAN-T5-Large & 500 & 12 & 20 & 4 & 718 & 67484 \\
StackOverflow & FLAN-T5-Large & 500 & 10 & 10 & 124 & 632 & 97037 \\
\bottomrule[\heavyrulewidth]
\end{tabular}
\vspace{-0.2cm}
\caption{More details on the datasets in our experiments. Each task besides TOFU has a minimum of 2 labels (TOFU is a next-token prediction task).}
\label{tab:datasets-full}
\end{table*}

\section{Additional experiments}
\label{app:experiments}
\textbf{Model selection and data formatting matters.} On both Reddit (Table~\ref{tab:cls_vs_nwp_reddit}) and Sent140 (Table~\ref{tab:cls_vs_nwp_sent140}), we find that model size is not necessarily the most important factor for good merging. In particular, it is important to consider (1) model architecture and (2) modeling format. With the proper modeling format, smaller models can outperform larger architectures from other model families. Generally, we see that using (1) an instruction-tuned model, (2) a language modeling rather than classification objective and (3) a prompt that achieves good zero-shot performance can greatly improve the performance of merging. ``CLS Head'' means we finetune a classification head, ``ZS Init'' means we initialize the classification heads using an embedding of the label text, ``LM Head'' means we keep the original language modeling head and finetune the model to output the label text, and ``Prompt'' means we design a prompt that encourages the model to output the label texts.

\begin{figure*}[]
    \centering    \includegraphics[width=0.7\linewidth]{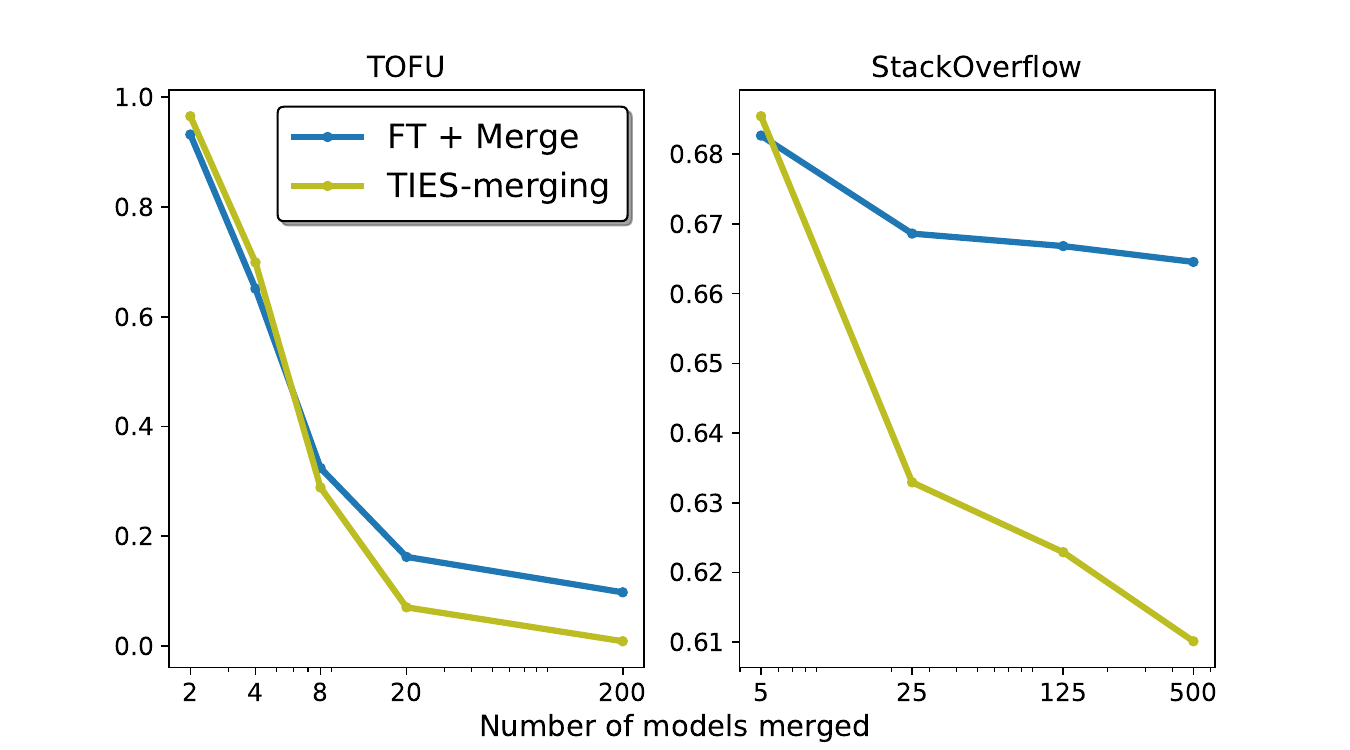}
    \caption{We compare \ties~to~\ft~while varying the number of models merged. Scale can disproportionately harm \ties~compared to regular \ft.}
    \label{fig:ties-tasks}
\end{figure*}

\begin{figure*}[]
    \centering    \includegraphics[width=1.0\linewidth]{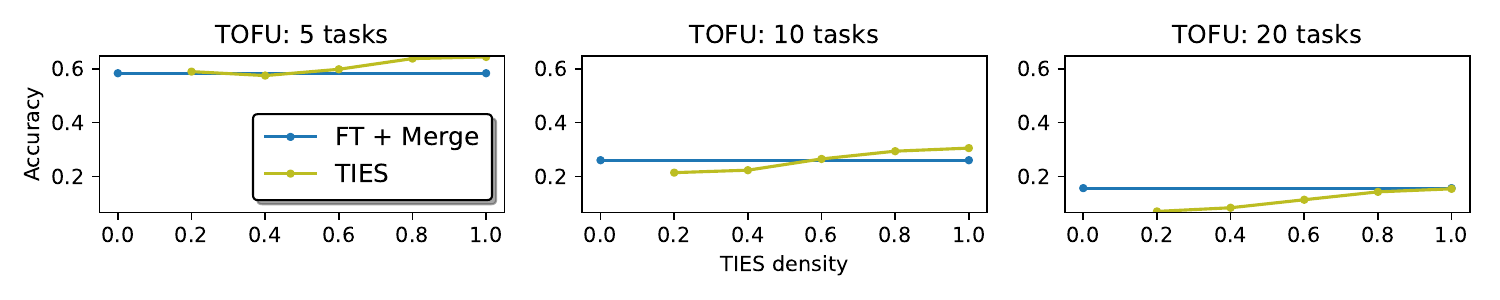}
    \includegraphics[width=1.0\linewidth]{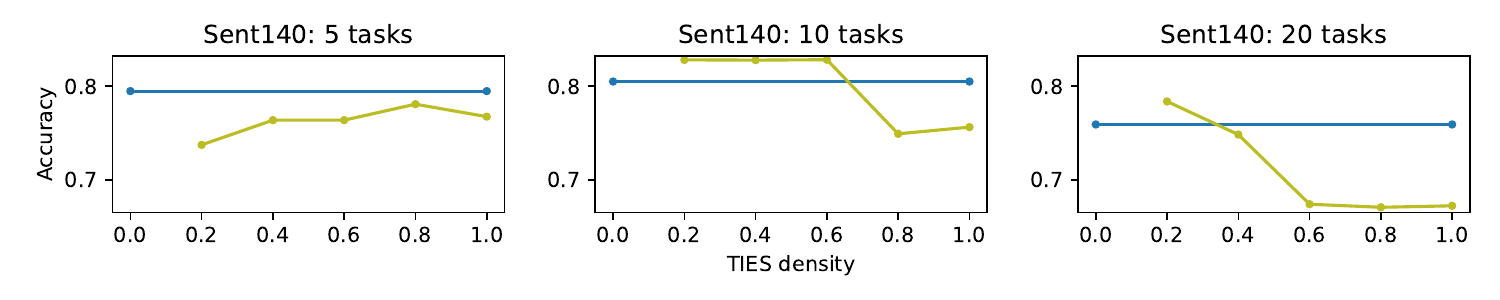}
    \includegraphics[width=1.0\linewidth]{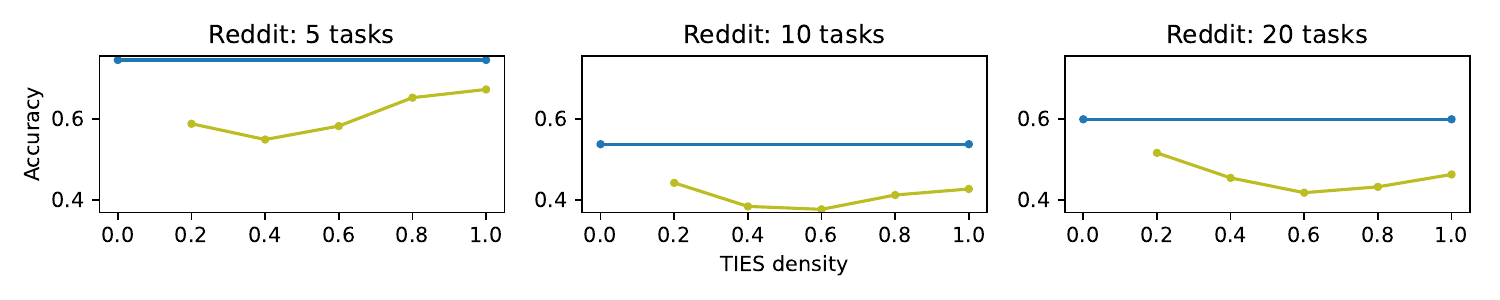}
    \includegraphics[width=1.0\linewidth]{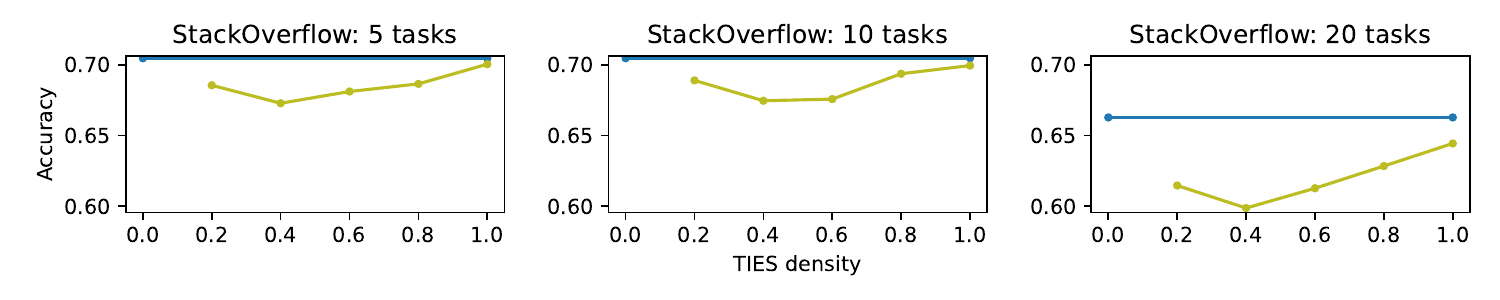}
    \caption{We compare \ties~to~\ft~while varying the density of \ties~and the number of models merged. First, sparsity (trimming) tends to be more beneficial when merging a larger number of models. However, in most settings, sparsity harms the accuracy of individual models and outweighs the benefits of reducing sign conflicts.}
    \label{fig:ties-density}
\end{figure*}

\textbf{\ties ablations.} As discussed in Section~\ref{sec:method}, \ties is a model merging method designed to reduce sign conflicts. It first ``trims'' i.e. prunes task vectors by parameter magnitude given a density hyperparameter, then ``elects'' a global sign vector based on the majority sign of each parameter. Finally, it merges the task vectors while only keeping the parameters which agree with the global sign. In Figure~\ref{fig:ties-tasks}, we show that \ties outperforms \ft at a small number of tasks (4-5), but degrades as the number of tasks increases i.e. more models are merged. In Figure~\ref{fig:ties-density}, we perform an ablation over the trimming density of \ties and the number of tasks merged. We find that trimming harms performance when there are only a few tasks, but can yield substantial improvements when there are more tasks.

\textbf{Systematic Evaluation.} In Table~\ref{tab:model_data}, we show a systematic evaluation over several model and dataset pairs. Across all settings, \ours outperforms all other baselines. In our main experiments, there were two key considerations for selecting the final model-dataset pairs. First, we selected base models with the best zero-shot performance on a particular dataset. This is a common heuristic for selecting models to finetune, particularly in model merging settings.
Second, since Sent140 is part of the FLAN dataset, we used Llama3.2 instead of FLAN-T5 for Sent140.

\textbf{Evaluation on MUSE-News.} We perform an additional evaluation which finetunes Llama2-7b-hf on MUSE-News~\cite{shimuse}, a collection of BBC News articles. We then unlearn the forget set of articles and  evaluate the model on a forget and retain set of QA pairs. To train models with SIFT-masks, we train one model for each article (task), which enables article-level unlearning. Since the relevant article is not provided with the query, we consider a simple nearest-neighbor bag-of-words baseline to select the most relevant article and apply its model to the query. We finetuned the model with LoRA rank 4.

We evaluate 4 metrics from the main results in the MUSE paper. Lower VerbMem (verbatim memorization) and Forget (forget set utility) is better, PrivLeak (how easily forget set membership can be inferred from an approximately vs exactly unlearned model; see C3. of the MUSE set for details) should be close to 0, and higher Retain (retain set utility) is better.

As expected, even though approximate unlearning methods (GA, NPO) are effective at lowering VerbMem and Forget, they lack fine control over unlearning and thus have high PrivLeak. Since merging and localization methods (\ft, \tall, \ours) modify the learning and unlearning algorithms such that unlearning is exact i.e. $f_{\text{retrain}} = f_{\text{unlearn}}$, the PrivLeak from these methods is zero. Compared to other unlearning baselines, \ours achieves the best Retain utility by a significant margin (13 points of ROUGE over \tall, 8 points over NPO\_GDR).

\textbf{Biased Rademacher.} Instead of uniformly sampling sign mask entries from $\{-1, +1\}$, we consider sampling from a biased distribution (i.e. $+1$ with probability $p$, and $-1$ with probability $1-p$). We evaluated this method on TOFU in Table~\ref{tab:biased_rademacher} and find that model utility remains similar.

\section{\ours pseudocode}
% \RestyleAlgo{boxed}
\RestyleAlgo{boxruled}
\LinesNumbered
\IncMargin{1em}
\SetAlCapHSkip{1em}
% \SetAlCapSkip{3em} % Adjust the value (e.g., 0.5em, 10pt) as needed
\begin{algorithm}[]
\caption{\ours~pseudocode}
\label{alg:sift}
Require: \\
$T$ (task indices) \\
$\{c_t\}_{t\in T}$ (tasks)\\
$M_0$ (pretrained model)\\
$E_{\text{tune}}$ (finetuning steps)\\
$\eta$ (learning rate) \\
$\mathcal{L}$ (loss function) \\
\SetKwFunction{FnSIFT}{SIFT}
\SetKwFunction{FnSIFTMASKS}{SIFT-masks}
\SetKwFunction{FnMerge}{Merge}
\SetKwFunction{FnUnmerge}{Unmerge}
\SetKwFunction{FnLocalize}{Localize}
\SetKwProg{Fn}{Function}{:}{}

\Fn{\FnSIFT{t, v}}{
    $\tau_t\gets \vec{0}$ \\
    $\texttt{opt}\gets\texttt{adam\_optimizer}(\tau_t, \eta)$\\
    \For{$i=1..E_{\text{tune}}$}{
        $M_t = M_0 + \tau_t$ \\
        $x,y\gets$ sample batch of data from $c_t$ \\
        % $\texttt{loss}\gets\texttt{cross\_entropy}(M_t(x), y)$ \\
        % $\texttt{loss.backward()}$ \\
        \# Compute gradient and update $\tau_t$ \\
        $\tau_t\texttt{.grad} \gets \nabla_{\tau_t} \mathcal{L}(M_t(x), y)$ \\
        % Backpropagate loss and update $\tau_i$ \\
        $\texttt{opt.step()}$ \\
        \# Apply sign mask \\
        $m_t \gets \mathbbm{1}\{\tau_t \odot v \geq 0\}$ \\
        $\tau_t \gets \tau_t \odot m_t$ \\
    }
    \KwRet $\tau_t, m_t$
}
\Fn{\FnMerge{$\{\tau_t\}_{t\in T}$}}{
    $\overline{\tau}\gets\sum_{t\in T}\tau_t$ \\
    \KwRet $\overline{\tau}$
}
\Fn{\FnUnmerge{$\overline{\tau}, t$}}{
    $\overline{\tau}\gets \tau - \tau_t$ \\
    $T \gets T \setminus \{t\}$ \\
    \KwRet $\overline{\tau}$ 
}
\Fn{\FnLocalize{$\overline{\tau}, t$}}{
    \KwRet $(\overline{\tau} \odot m_t) / |T|$ 
}
\Fn{\FnSIFTMASKS{$T$}}{
    $v\gets\text{vector where } v_i \in \{+1,-1\}$ uniformly i.i.d.\\
    \For{$t=1..T$}{
        $\tau_t, m_t = \FnSIFT(t, v)$ \\
    }
    $\overline{\tau}\gets\FnMerge(\{\tau_t\}_{t\in T})$ \\
    \KwRet $\overline{\tau}$ 
}
\end{algorithm}
\DecMargin{1em}

\begin{table*}[]
\normalsize
\centering
\begin{tabular}{r|ccccc}
\toprule[\heavyrulewidth]
 & \multicolumn{5}{c}{Sent140} \\
\midrule
Model & Params & Format & Zeroshot & Finetune & Merge \\
\midrule
GPT2-XL & 1.61B & CLS Head & 50.0 & 85.7 & 70.8 \\
\midrule
Llama 3.2-1B & 1.24B
  & CLS Head & 50.0 & 69.4 & 55.3 \\
& & CLS Head + ZS Init & 50.0 & 67.4 & 53.0 \\
& & LM Head + Prompt & 68.5 & 85.5 & \textbf{77.1}  \\
% \midrule
% Global & 71.6 & 77.7 & \multicolumn{2}{|c}{N/A} \\
\bottomrule[\heavyrulewidth]
\end{tabular}
\caption{Merging 100 clients on Sent140. GPT2-XL and Llama 3.2 can achieve similar local accuracy after finetuning, but Llama 3.2 has much better accuracy after merging.}
\label{tab:cls_vs_nwp_sent140}
\end{table*}
\begin{table*}[]
\normalsize
\centering
\begin{tabular}{r|ccccc}
\toprule[\heavyrulewidth]
 & \multicolumn{5}{c}{Reddit} \\
\midrule
Model & Params & Format & Zeroshot & Finetune & Merge \\
\midrule
% Llama 3.2-1B & 1.24B
%   & CLS Head & 50.0 & 69.4 & 55.3 \\
% & & CLS Head + ZS Init & 50.0 & 67.4 & 53.0 \\
Llama 3.2-1B & 1.24B
& LM Head + Prompt & 10.0 & 75.7 &  24.9 \\
\midrule
FLAN-T5-Large & 783M
  & CLS Head & 3.0 & 61.3 & 19.3 \\
& & CLS Head + ZS Init & 7.1 & 61.8 & 16.5 \\
& & CLS Head + ZS Init + Prompt & 6.7 & 70.5 & 21.7\\
& & LM Head + Prompt & 29.7 & 76.5 & \textbf{43.2} \\
\bottomrule[\heavyrulewidth]
\end{tabular}
\caption{Merging 100 clients on Reddit. While Llama 3.2 does well on Sent140, it does poorly on Reddit, indicating that testing multiple model architectures can be helpful. We do not test FLAN-T5 on Sent140 because Sent140 is included in the FLAN dataset. Unlike the Reddit setting presenting in the main paper, here we ran experiments on a setting where each client's data is subsampled to only hold 50 examples from each of its 2 most common labels.}
\label{tab:cls_vs_nwp_reddit}
\end{table*}

% \begin{figure*}[h!]
%     \centering    \includegraphics[width=0.7\linewidth]{figures/8-ties-c.pdf}
%     \caption{We compare \ties~to~\ft~and \texttt{SI\ft} when merging 500 models. While both \ties~and \texttt{SI\ft} attempt to reduce sign conflicts, this does not significantly affect merging quality.}
%     \label{fig:ties-merge}
% \end{figure*}

\begin{table*}[]
\normalsize
\centering
\caption{Performance comparison across different methods and datasets}
\label{tab:model_data}
\begin{tabular}{lcccc}
\midrule
\textbf{Model} & \textbf{Zeroshot} & \textbf{\ft} & \textbf{\tall} & \textbf{\ours} \\
\midrule
\multicolumn{5}{c}{\textbf{TOFU}} \\
\midrule
GPT2-XL & \textbf{11.1} & 11.1 & 50.3 & \textbf{95.7} \\
FLAN-T5-Large & 8.1 & 8.0 & 18.7 & 47.5 \\
Llama3-1B-Instruct & 9.8 & 9.7 & 41.2 & 82.4 \\
\midrule
\multicolumn{5}{c}{\textbf{Sent140}} \\
\midrule
GPT2-XL & 0.0 & 62.5 & 72.3 & 70.0 \\
FLAN-T5-Large & \textbf{82.7} & 84.0 & 85.8 & \textbf{86.3} \\
Llama3-1B-Instruct & 73.4 & 78.3 & 82.5 & 83.8 \\
\midrule
\multicolumn{5}{c}{\textbf{Reddit}} \\
\midrule
GPT2-XL & 0.0 & 20.0 & 35.6 & 68.4 \\
FLAN-T5-Large & \textbf{46.2} & 61.1 & 74.8 & \textbf{78.7} \\
Llama3-1B-Instruct & 0.0 & 25.0 & 60.3 & 51.8 \\
\midrule
\multicolumn{5}{c}{\textbf{StackOverflow}} \\
\midrule
GPT2-XL & 0.1 & 62.1 & 76.6 & 76.5 \\
FLAN-T5-Large & \textbf{48.8} & 64.8 & 69.7 & 71.2 \\
Llama3-1B-Instruct & 34.7 & 81.3 & 85.9 & \textbf{87.4} \\
\midrule
\end{tabular}
\end{table*}

\begin{table*}[h]
\normalsize
\centering
\caption{Evaluation on MUSE-News}
\label{tab:muse-news}
\begin{tabular}{lcccc}
\midrule
\textbf{Method} & \textbf{VerbMem} & \textbf{Forget} & \textbf{PrivLeak} & \textbf{Retain} \\
\midrule
Target (pre-unlearn) & 58.4 & 63.9 & -99.8 & 55.2 \\
Retrain (unlearned) & 20.8 & 33.1 & 0 & 55.0 \\
GA\_GDR & 4.9 & 31.0 & 108.1 & 27.3 \\
NPO\_GDR & 1.2 & 54.6 & 105.8 & 40.5 \\
\ft & 18.7 & 30.8 & 0 & 33.1 \\
\tall & 19.1 & 29.3 & 0 & 35.1 \\
\ours & 18.8 & 27.0 & 0 & 48.6 \\
\midrule
\end{tabular}
\end{table*}

\begin{table*}[h]
\normalsize
\centering
\caption{Performance comparison across different LoRA rank}
\label{tab:lora_rank}
\begin{tabular}{lcccc}
\toprule
\textbf{Reddit} & \textbf{LoRA (r=4)} & \textbf{LoRA (r=16)} & \textbf{LoRA (r=64)} & \textbf{Full FT} \\
\midrule
\tall & 73.7 & 73.8 & 74.7 & 74.2 \\
\ours & 76.4 & 76.5 & 76.8 & 80.2 \\
\bottomrule
\end{tabular}
\vspace{1em}
\begin{tabular}{lcccc}
\toprule
\textbf{StackOverflow} & \textbf{LoRA (r=4)} & \textbf{LoRA (r=16)} & \textbf{LoRA (r=64)} & \textbf{Full FT} \\
\midrule
\tall & 66.2 & 69.4 & 71.9 & 72.5 \\
\ours & 66.7 & 70.5 & 73.2 & 76.2 \\
\bottomrule
\end{tabular}
\end{table*}

\begin{table*}[h]
\normalsize
\caption{Biased Rademacher Evaluation}
\label{tab:biased_rademacher}
\centering
\begin{tabular}{lccc}
\toprule
Biased Rademacher $p$ & 0.5 (default) & 0.7 & 0.9 \\
\midrule
TOFU Probability & 0.965 & 0.963 & 0.962 \\
\bottomrule
\end{tabular}
\end{table*}

\clearpage

\end{document}